\documentclass{article} %
\usepackage[preprint]{colm2025_conference}

\usepackage{microtype}
\usepackage{hyperref}
\usepackage{url}
\usepackage{booktabs}

\usepackage{lineno}
\usepackage{graphicx} 

\usepackage{fvextra}
\usepackage{subcaption}

\usepackage{rotating}  
\usepackage{enumitem}

\DefineVerbatimEnvironment{verbatim}{Verbatim}{breaklines=true}

\definecolor{darkblue}{rgb}{0, 0, 0.5}
\hypersetup{colorlinks=true, citecolor=darkblue, linkcolor=darkblue, urlcolor=darkblue}
\usepackage{xspace}
\usepackage[T1]{fontenc}  %
\usepackage[most]{tcolorbox} %
\tcbuselibrary{listings}     %

\newtcblisting{promptbox}[1][]{
  colback=white,              %
  colframe=black,             %
  fonttitle=\bfseries,  
  title=#1,                   %
  arc=2mm,                    %
  listing only,               %
  listing options={
    basicstyle=\ttfamily,     %
    columns=fullflexible,     %
    frame=none,               %
    breaklines=true,          %
    tabsize=1                %
  }
}
\newcommand{\benchmark}{\textsc{LogiPlan}\xspace}

\title{LogiPlan: A Structured Benchmark for Logical Planning and Relational Reasoning in LLMs}

\author{
Yanan Cai\thanks{\texttt{yanacai@microsoft.com}} \\Microsoft Azure  \And 
Ahmed Salem\thanks{\texttt{ahmsalem@microsoft.com}} \\Microsoft  \And
Besmira Nushi\\Microsoft Research \And
Mark Russinovich\\Microsoft Azure
}

\newcommand{\mypara}[1]{\noindent{\bf {#1}.}}

\newcommand{\gen}{Plan Generation\xspace}
\newcommand{\cons}{Consistency Detection\xspace}
\newcommand{\qa}{Comparison Question\xspace}

\newcommand{\ClaudeSonnetThinking}{{Claude 3.7 Sonnet}\xspace}  
\newcommand{\ROne}{{DeepSeek R1}\xspace} 
\newcommand{\GeminiPro}{{Gemini 2.0 Pro}\xspace}  
\newcommand{\GeminiFlash}{{Gemini 2 Flash Thinking}\xspace}  
\newcommand{\LlamaThreeOneLarge}{{Llama 3.1 405B}\xspace} 
\newcommand{\GPTFourO}{{GPT-4o}\xspace}  
\newcommand{\GPTFourFive}{{GPT-4.5}\xspace}  
\newcommand{\OOne}{{O1}\xspace}  
\newcommand{\OThree}{{O3-mini}\xspace} 

\begin{document}

\ifcolmsubmission
\linenumbers
\fi

\maketitle

\begin{abstract}
We introduce \benchmark, a novel benchmark designed to evaluate the capabilities of large language models (LLMs) in logical planning and reasoning over complex relational structures. Logical relational reasoning is important for applications that may rely on LLMs to generate and query structured graphs of relations such as network infrastructure, knowledge bases, or business process schema. Our framework allows for dynamic variation of task complexity by controlling the number of objects, relations, and the minimum depth of relational chains, providing a fine-grained assessment of model performance across difficulty levels.
\benchmark encompasses three complementary tasks: (1) \emph{\gen}, where models must construct valid directed relational graphs meeting specified structural constraints; (2) \emph{\cons}, testing models' ability to identify inconsistencies in relational structures; and (3) \emph{\qa}, evaluating models' capacity to determine the validity of queried relationships within a given graph. Additionally, we assess models' self-correction capabilities by prompting them to verify and refine their initial solutions.
We evaluate state-of-the-art models including \ROne, \GeminiPro, \GeminiFlash, \GPTFourFive, \GPTFourO, \LlamaThreeOneLarge, \OThree, \OOne, and \ClaudeSonnetThinking across these tasks, revealing significant performance gaps that correlate with model scale and architecture. Our analysis demonstrates that while recent reasoning-enhanced models show promising results on simpler instances, they struggle with more complex configurations requiring deeper logical planning. 
\end{abstract}
\section{Introduction}
Recent advances in large language models (LLMs) have sparked interest in their ability to perform complex reasoning tasks~\citep{guo2025deepseek,O3mini,jaech2024openai}. While 
current work showcases increased reasoning skills in math and coding skills, there are several real-world problems which require reasoning skills that are more algorithmic in nature and often computationally expensive. We still lack an in-depth understanding of how current models perform and how their performance scales with the size of these problems. Logical relational reasoning is a problem in this category, which requires models to plan or validate relationships between a large number of objects. For example, in operations research and business planning for supply chains~\citep{tsouros2023holy,li2023large}, there exist several relational planning problems in which each object is a task or an event, and a relationship is a constraint that specifies whether a task should happen before or after another task. Similar examples also exist in code analysis~\citep{Neamtiu2005UnderstandingSC}, and agentic system~\citep{wu2023autogen, fourney2024magentic} where dependencies are introduced and task planning and orchestration is required. The rapid integration of LLMs chatbots and agentic AI systems has transformed how people interact with and manage knowledge through natural language. This evolution necessitates more rigorous evaluation of relational reasoning to better understand the advantages and limitations of LLMs.

In this work, we introduce \benchmark, the first benchmark specifically designed to assess both the generation of structured relational plans and the subsequent reasoning about these plans. Our framework fills an important gap by focusing on the entire pipeline—from the autonomous construction of relational graphs to the critical examination of their logical consistency—which is central to many real-world applications.

The structured form of \benchmark enables dynamically varying parameters like the number of objects, comparative relations, and the minimum depth of relational chains. This facilitates a fine-grained analysis of model performance across a wide spectrum of difficulties. This approach mirrors the challenges faced by intelligent agents operating in unpredictable environments, where the ability to generate and validate continuously updated plans is critical for successful deployment. Dynamic data generation also offers the advantage that it allows for generating fresh or private copies of the benchmark with controlled difficulty, in cases where benchmark memorization becomes a concern. 

\benchmark consists of three complementary tasks:  (1) \emph{\gen}, where models must construct valid directed graphs with comparative relationships meeting specified structural constraints. (2) \emph{\cons}-Does a given graph have inconsistencies or cycles? What are they?; and (3) \emph{\qa}-Is a relationship valid for a given graph?. Through a set of \benchmark experiments, we present a comprehensive evaluation of nine state-of-the-art models, encompassing both instruction-tuned and reasoning-capable LLMs. Our analysis uncovers significant performance disparities that correlate with model scale and architecture, providing valuable insights into the evolving capabilities of LLMs.

Our evaluation reveals a significant gap between the reasoning-based and instruction-based models in the \gen task. This difference stems from the reasoning models' ability to identify a straightforward generation technique where objects maintain logical relationships, such as $A>B>\cdots>Z$. For the \cons and \qa tasks, the gap persists, primarily between \OOne and \OThree and the other models. In particular, \cons remains the most difficult task, where even state-of-the-art reasoning models face challenges as the size of the problem increases. Specifically, the performance drop for this task happens much earlier for models like \ROne and \GeminiFlash.

Overall, \benchmark serves as both a diagnostic tool and a catalyst for future research, pushing the boundaries of what is achievable in logical relational reasoning with LLMs. We will open source the benchmark, the code for data generation, and the evaluation to enable future research on LLM evaluation and reasoning.

\section{\benchmark Construction and Data Generation}

Our approach focuses on developing a dynamic benchmark to rigorously probe the logical planning and relational reasoning capabilities of LLMs. The methodology underlying \benchmark is designed to ensure comprehensive coverage of the targeted reasoning tasks while maintaining control over the complexity of the generated relational structures.

\subsection{Data Synthesis}
We adopt a data generation strategy that creates synthetic relational graphs with controllable complexity. By adjusting parameters, such as the number of objects, relations, and the minimum depth of relational chains, we synthesize tasks that systematically probe the models’ capabilities. This approach allows us to generate diverse and scalable problem instances, ensuring that the benchmark reflects both simple and highly complex reasoning scenarios.

\subsubsection{\gen}

The \gen task is motivated by real-world applications where models must autonomously construct complex relational structures—such as network topologies or business process models—without human intervention. The primary goal is to evaluate the models’ ability to produce consistent, accurate, and complete plans that adhere to specified constraints.

\mypara{Task Design}
In this task, each model generates a relational graph based on given specifications (i.e., number of objects and relations). The evaluation encompasses several dimensions: the overall accuracy in generating the required relations, the consistency of the constructed graph in terms of logical coherence, the absence of redundant relations. This rigorous evaluation framework ensures that the models are not only capable of constructing a plan but also of maintaining structural integrity and coherence as the complexity of the task increases.

\mypara{Data Generation}
We generated datasets of varying complexity using a Python program that systematically pairs the number of object with the number of relations. The dataset comprises pairs with object counts ranging from 3 to 50. For each object configuration, we varied the number of relations from minimal (equal to the object count $n$) to dense (approaching $n(n-1)/2$, with practical upper limits), creating a diverse testbed for evaluating language model performance on graph-structured generation tasks.

\subsubsection{\cons}

In many applications, especially those involving critical infrastructure or knowledge management, identifying inconsistencies such as cycles or contradictory relations is important. The \cons task is designed to assess a model’s ability to scrutinize a given graph and pinpoint logical errors. Note that the algorithmic complexity of cycle detection in a directed graph is $O((V+E)×(C+1))$ as shown by \cite{johnson1975finding}, where V is the number of vertices, E is the number of edges, C is the total number of simple cycles in the graph.

\mypara{Task Design} For this task, models are presented with relational graphs that may contain cycles or contradictions. The evaluation focuses on two main aspects: the model's ability to detect the presence of inconsistencies and its ability to correctly identify and extract these inconsistencies. This involves assessing the effectiveness in detecting true inconsistencies while minimizing false alarms, and providing a balanced measure of its detection capabilities. By varying the number of objects, relations, and the minimum cycle length (how many steps are needed for the cycle to close), we challenge the models with increasingly complex scenarios, ensuring a comprehensive assessment of their verification capabilities.

\mypara{Data Generation} We used the Python NetworkX library to construct controlled relational graphs, encompassing both positive cases (acyclic, consistent graphs) and negative cases (cyclic, inconsistent graphs). The algorithms generate directed graphs with parameterized complexity, varying the number of objects and the density of relations while ensuring complete connectivity. For cyclic cases, we enforce minimum cycle lengths in the graph to assess multi-step reasoning abilities. These graphs are then transformed into randomized textual statements where: (1) objects receive arbitrary labels unrelated to their structural positions, (2) relations are inconsistently expressed using both forward and backward notation (randomly choosing between $>$ or $<$), and (3) statements are deliberately presented in shuffled order. This intentional randomization prevents models from solving tasks through superficial pattern recognition, instead requiring logical reasoning to reconstruct the complete relation network and identify cycles regardless of their surface representation. As an example, \autoref{fig:graph-example} in Appendix illustrates an example of the graph used in the benchmark dataset. 

\subsubsection{\qa}
The \qa task simulates scenarios where models must interpret a pre-existing relational graph and make informed judgments about specific relations. This task is pivotal for applications such as automated reasoning in databases or dynamic knowledge base querying. Note that the algorithmic complexity of the reachability check between two nodes in a directed graph is $O(V+E)$ using Breadth-First Search (BFS) or Depth-First Search (DFS).

\mypara{Task Design} In this task, models are required to evaluate relational statements and categorize them as ``True'', ``False'', or ``Unknown''. The Unknown category is particularly significant as it assesses a model’s ability to handle incomplete information, i.e., situations where the given graph is insufficient to answer the query about the specified relation. We challenge the models with increasingly complex problems by examining performance based on the number of objects, relations, and the required inference depth. This design tests not only direct reasoning capabilities but also the model's decision-making process in complex and edge cases.

\mypara{Data Generation}
We use the same graph generation algorithm as the \cons task. For each graph, we randomly select object pairs at specific path distances and randomly determine whether to present them as true or false comparisons. A comparison is marked as ``True'' when it accurately reflects the graph relationship, and ``False'' when it contradicts it. ``Unknown'' pairs are specifically created from objects with no connecting paths.

\subsubsection{Self Correction}
In our benchmark, we also evaluate the models' self-correction capabilities. A model's ability to self-assess and correct its output when necessary can demonstrate a logical trait, ensuring that it does not merely alter its responses when questioned by the user.

\mypara{Task Design} Following the completion of the \cons and \qa tasks, models are given a follow-up prompt: \emph{Are you sure?}. We assess two key aspects: the degree to which initial decisions are revised and the impact of these revisions on overall accuracy. Specifically, we examine whether self-correction leads to improved outcomes or introduces additional errors. This evaluation sheds light on how internal confidence calibration mechanisms influence the decision-making process, highlighting the model's capability to adapt and refine its outputs in response to uncertainty.

\section{Evaluation}

In this section, we present a comprehensive evaluation of our \benchmark framework across a diverse array of state-of-the-art models.

\subsection{Metrics}

To rigorously evaluate the performance of different models on our tasks, we employ a diverse set of metrics tailored to each task's unique requirements. This allows us to deeply investigate the models' performance across varying computational demands.

\mypara{\gen} For the \gen task, we primarily use \emph{Accuracy}, which measures the overall success rate of the model in generating correct plans. A perfect accuracy score (100\%) implies that the model correctly generates the required number of relations with the correct number of objects, and without any inconsistencies. 

Specifically, we use more granular metrics to determine the overall \emph{Accuracy}:

\begin{itemize}[leftmargin=*, itemsep=0em]
    \item\emph{Consistency}: This metric evaluates whether the generated relations are free from logical inconsistencies. A consistency score of 100\% indicates no inconsistencies.

    \item\emph{Duplicates}: This measures the presence of duplicate relations. A score of 100\% means no duplicates are detected.

    \item\emph{Object Numbers} and \emph{Relation Numbers}: These metrics assess whether the generated objects and relations match the specified numbers in the task. A perfect score of 100\% indicates full compliance with the task requirements.
\end{itemize}

\mypara{\cons} For the \cons task, we use \emph{F1 score} calculated from \emph{Precision} and \emph{Recall} to assess the correctness of the model's decisions. We evaluate two aspects: (1) whether the model correctly identifies the presence or absence of inconsistencies, and (2) if inconsistencies are present, whether the model correctly identifies and outputs all of them. Specifically:
\begin{itemize}[leftmargin=*, itemsep=0em]
    \item\emph{Precision}: For consistency cases, \emph{Precision} measures the correctness of the model reporting consistency. For inconsistency cases, it measures the proportion of correctly identified inconsistencies among all inconsistencies reported by the model.
    
    \item\emph{Recall}: For consistency cases, \emph{Recall} is always 1 (as there is no inconsistency to find in the ground truth). For inconsistency cases, it measures the proportion of actual inconsistencies in the ground truth that were correctly identified by the model.
\end{itemize}

\mypara{\qa} For the \qa task, we measure \emph{Accuracy} to assess the model's ability to correctly determine the veracity of given relations. The potential model outputs are as follows:
\begin{itemize}[leftmargin=*, itemsep=0em]
    \item ``True'': Indicates that the comparison statement can be inferred from the graph.
    \item ``False'': Means the comparison can be inferred from the graph, but the statement is in the opposite direction.
    \item ``Unknown'': Implies that the relation between the two objects in the comparison statement cannot be determined from the graph. This option aims to evaluate the model's effectiveness in handling ambiguous or incomplete information.
\end{itemize}

\mypara{Self Correction} Finally, across the \cons and \qa tasks, we evaluate the model's ability to self-correct its output by measuring the change in its decisions when prompted with the query, ``Are you sure?''. This metric reveals whether the model's confidence adjustments lead to more accurate or erroneous outcomes, providing insights into its self-assessment capabilities.

\subsection{Evaluation Settings}

For all of our experiments (unless explicitly mentioned otherwise), we conduct five independent runs and report the average performance to account for the models' inherent non-deterministic nature and increase the robustness of our results.

We evaluate the following models under their respective settings. The detailed configuration for each model is presented in the Appendix (\autoref{tab:models}).

\mypara{Instruction-based models}  \LlamaThreeOneLarge, \GeminiPro, \GPTFourO, \GPTFourFive.

\mypara{Reasoning models} \OOne, \OThree, \ROne, \GeminiFlash, \ClaudeSonnetThinking.

\subsection{Results}

We now present the evaluation results for \benchmark's different tasks.

\subsubsection{\gen}

\begin{figure*}[!t]
\centering
\begin{subfigure}{0.48\columnwidth}
\centering
\includegraphics[width=1\columnwidth]{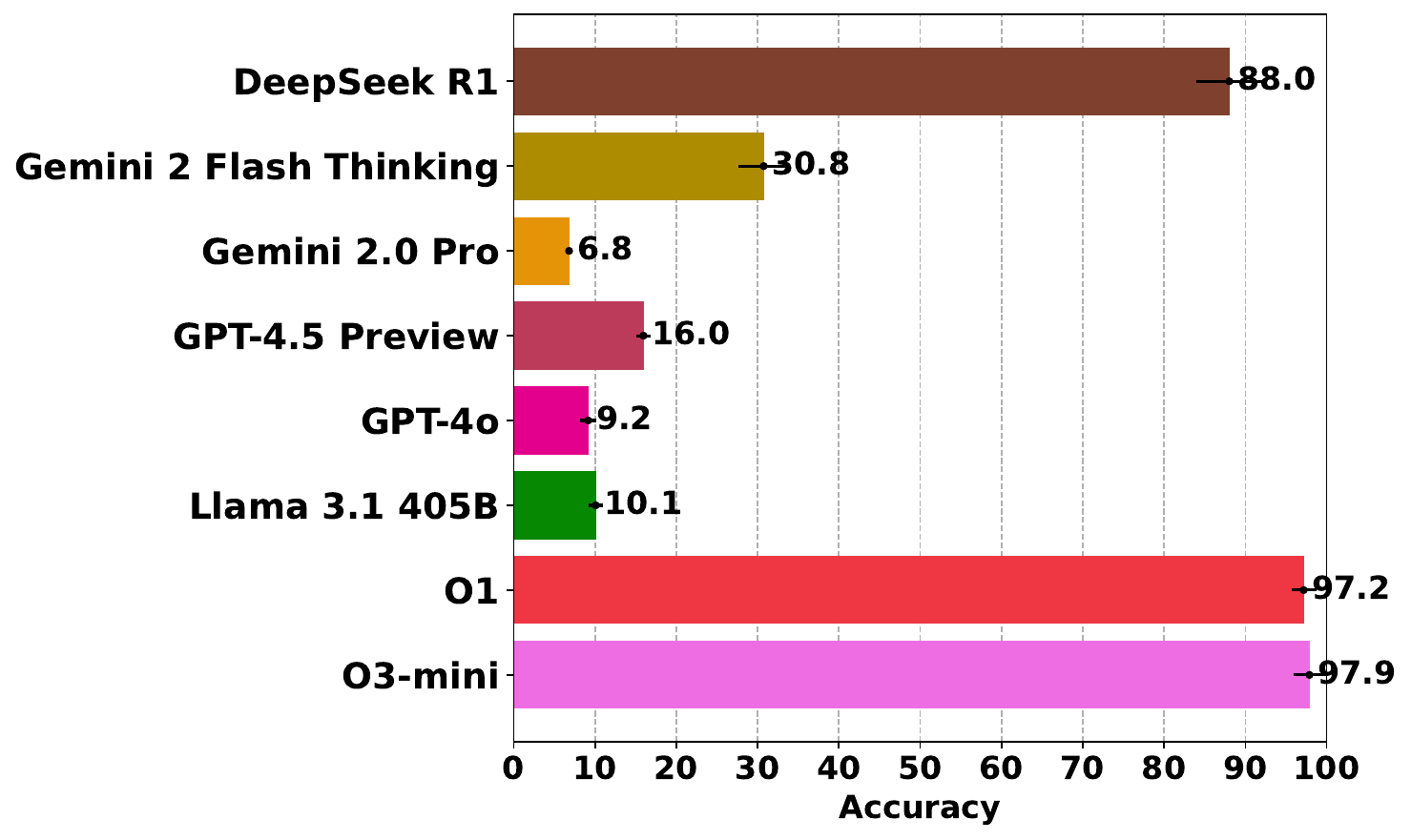}
\caption{Overall Accuracy.}
\label{fig:task1MainOverAll}
\end{subfigure}
\begin{subfigure}{0.48\columnwidth}
\centering
\includegraphics[width=1\columnwidth]{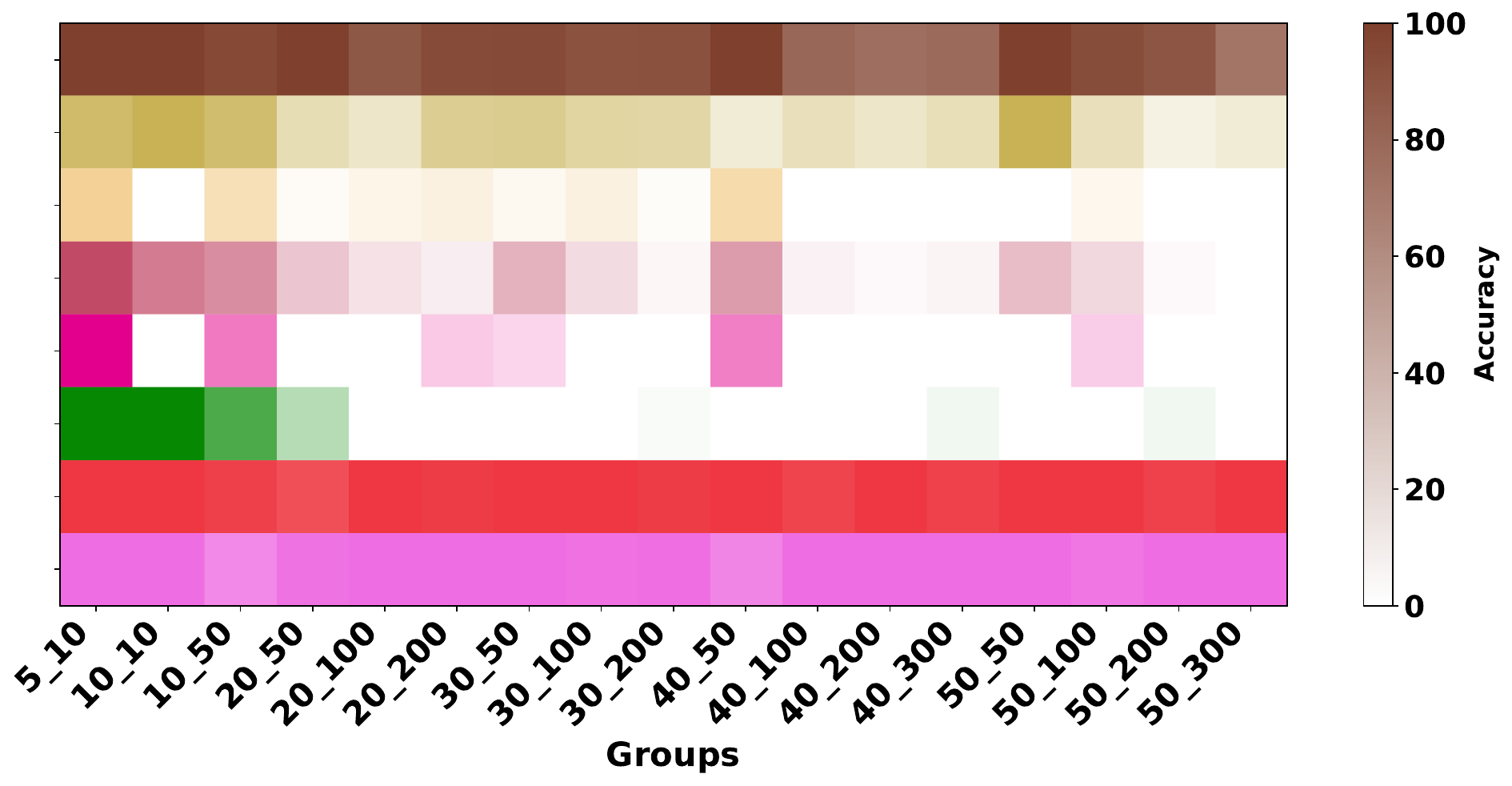}
\caption{Accuracy Per Group.}
\label{fig:task1MainFineGrained}
\end{subfigure}
\caption{Overview Accuracy of different models for the \gen task.}
\label{fig:task1Main}
\end{figure*}

We begin with the \gen task as a warm up task. In this task, each model generates a consistent list of relations based on the number of objects and relations specified in the prompt, evaluated over three independent runs. Our dataset comprises 142 examples, with objects ranging from 3 to 50 and relations ranging from 3 to 300. For each group of object numbers, the number of relations is incrementally increased from a minimum edge with a minimal step to a maximum reach evenly. Detailed configurations of these 142 examples and prompt template are provided in \autoref{append:detailedEval}.

We first present the overall accuracy (average across all runs) for each model, as shown in \autoref{fig:task1MainOverAll}. The figure highlights a substantial performance gap between reasoning models and instruction-based ones. For instance, the \OThree and \OOne models achieve notable accuracies of $97.9\%$ and $97.2\%$, respectively, while \ROne attains $88.0\%$. Interestingly, the \GeminiFlash model, despite being a reasoning model, only achieves $30.8\%$, which is significantly lower. However, it still performs approximately twice as well as the best instruction model \GPTFourFive. We further check the \GeminiFlash's failed cases, the main issue is its ability to avoid the presence of duplicates in its output.

Next, we present a more fine-grained analysis of accuracy by dividing the settings into bins and reporting each model's performance, as illustrated in \autoref{fig:task1MainFineGrained}. In this figure, the x-axis is labeled as $x\_y$ represents groups covering the number of objects between $(x_{i-1}, x_i]$ and the number of relations between $(y_{i-1}, y_i]$. This analysis reveals that reasoning models consistently exhibit strong performance, even as the number of relations and objects increases. Conversely, instruction-based models perform well only with limited numbers of relations and objects. In particular, when the number of objects is fixed, instruction-based models consistently perform worse as the number of relations increases.

Manual inspection of the reasoning models' outputs revealed a noticeable pattern; these models tend to employ a straightforward algorithm for ordering objects, such as $A > B > C > D$, and so on. This strategy allows the models to generate an arbitrary number of consistent relations. To further test their robustness, we provided a list of objects as random strings. Remarkably, the models maintained the simple ordering algorithm and produced consistent relations, demonstrating their robustness in maintaining logical consistency. Detailed plots are presented in \autoref{append:detailedEvalResult}. Albeit this is only one example, from an algorithm discovery perspective, the fact that models were able to figure out a straightforward solution for the generation, is an interesting indication that longer step-by-step traces can practically facilitate the discovery and execution of algorithmic approaches to problem solving.

\subsubsection{\cons}
\begin{figure*}[!t]
\centering
\begin{subfigure}{0.47\columnwidth}
\centering
\includegraphics[width=1\columnwidth]{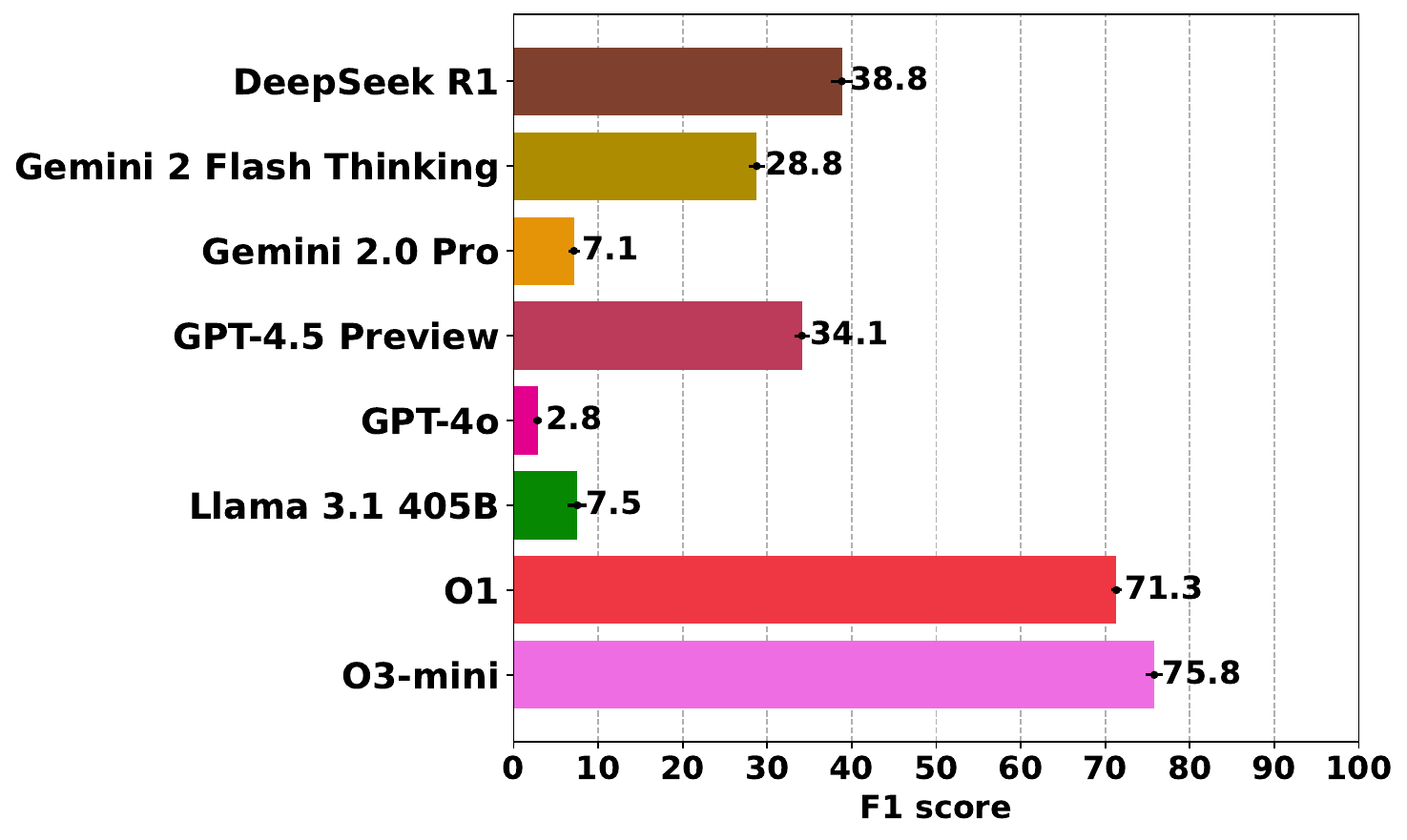}
\caption{Overall F1 Score.}
\label{fig:task2MainOverAll}
\end{subfigure}
\begin{subfigure}{0.52\columnwidth}
\centering
\includegraphics[width=1\columnwidth]{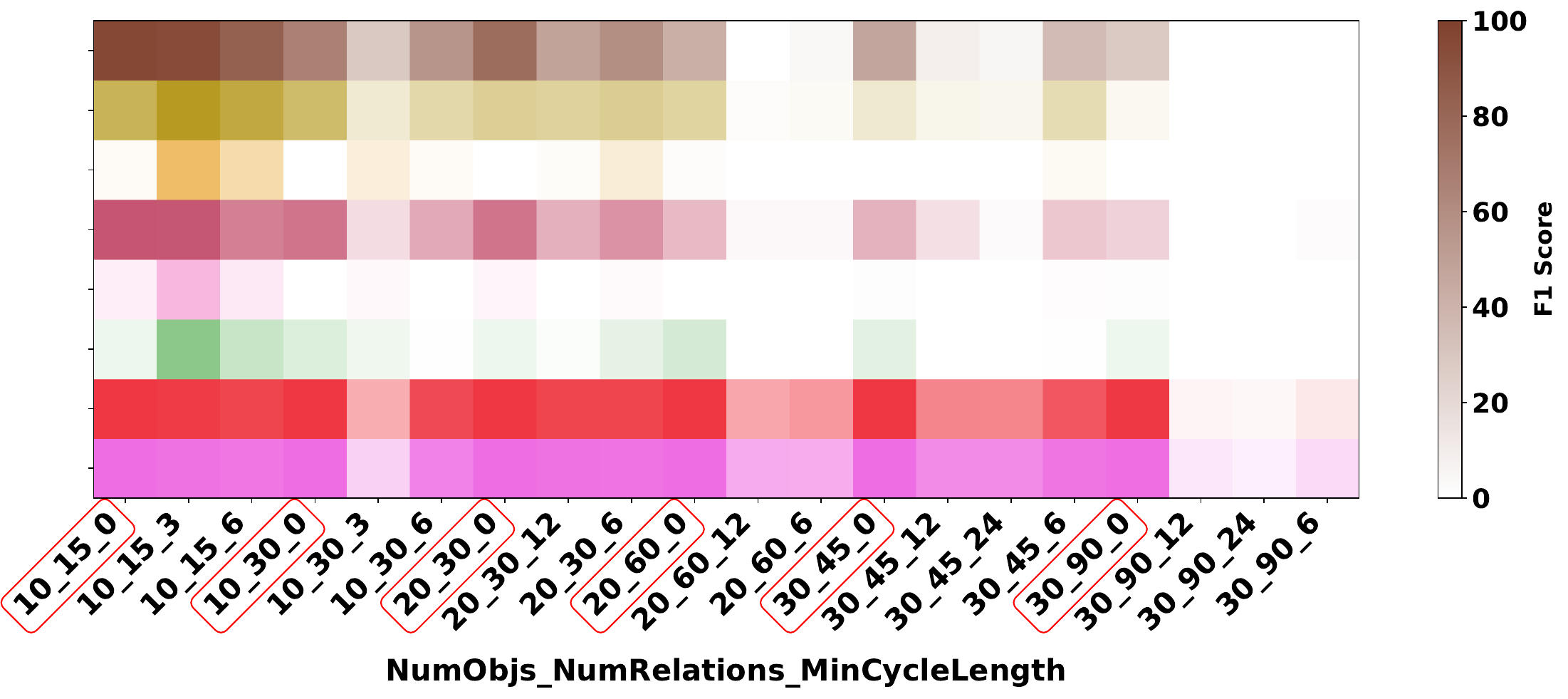}
\caption{F1 Score Per Group.}
\label{fig:task2MainFineGrained}
\end{subfigure}
\caption{Overview F1 Score of different models for the \cons task.}
\label{fig:task2Main}
\end{figure*}

Next, we evaluate the \cons task. For this task, we consider 20 different groups, each defined by varying numbers of objects, relations, and the minimum depth of cycles. For example, in a list containing cycles such as $A > B > C > A$, $A > B > A$, and $B > D > A > B$, the minimum cycle depth is 2, with objects A and B involved in the shortest cycle $A > B > A$. For consistent relations, we set the minimum cycle length as 0 in the group name. For each group, we generate 20 different plans, resulting in a total of 400 samples. For each sample, the model is tasked with detecting any cycles/contradictions and, if present, outputting all of them. More details on the data groups and the prompt template can be found in \autoref{append:detailedEval}.

Average F1 scores are shown in \autoref{fig:task2MainOverAll}. Unlike the \gen task, this task is more challenging for both reasoning and instruction-based models. The performance gap between the \OThree and \OOne models and the next best model (DeepSeek R1) is significant, almost 2x. Interestingly, in this task, GPT 4.5 achieves better performance than \GeminiFlash, indicating that an instruction-based model can outperform a reasoning model. However, the remaining instruction-based models perform significantly worse (by more than 3.8x) than the closest reasoning model.

Next, we report the fine-grained F1 scores for each model and configuration independently in \autoref{fig:task2MainFineGrained}. In this figure, the x-axis is labeled as $x\_y\_z$ represents groups covering the number of objects, the number of relations, and the minimum cycle length in the graphs. Note that every graph has a randomized numbers of cycles, with the enforcement on the minimum cycle length. The average number of cycles across all samples is 4.37. Similar to the \gen task, the larger the number of objects and relations, the more challenging the task becomes for the models. We also observe that longer cycles increase the difficulty for the models. This is evident when comparing settings with a minimum length of 0 (highlighted with red rectangles in the figure) to the two subsequent columns.

\begin{figure*}[!t]
\centering
\begin{subfigure}{0.49\columnwidth}
\centering
\includegraphics[width=1\columnwidth]{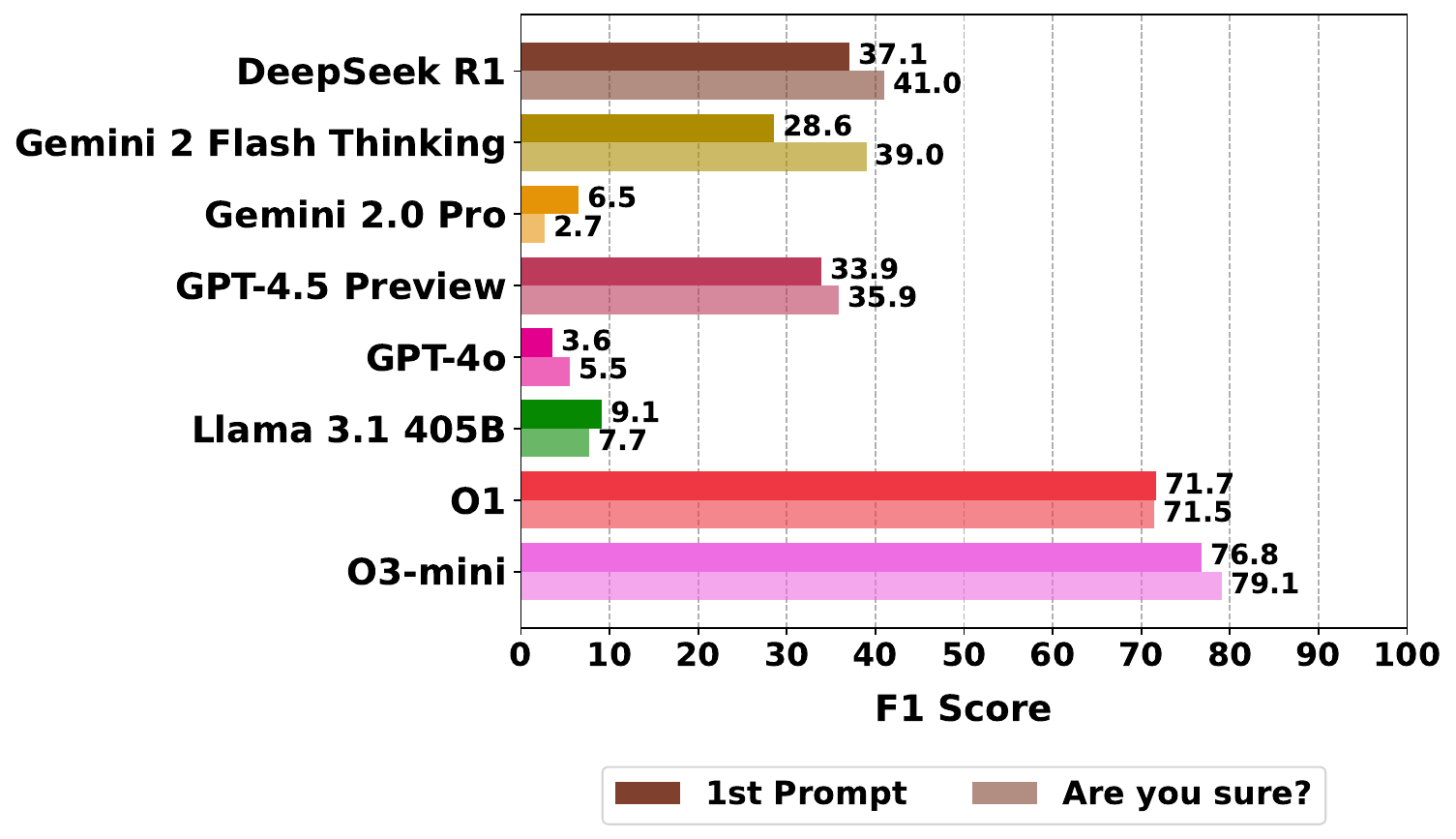}
\caption{Self Correction F1 Score.}
\label{fig:task2SC}
\end{subfigure}
\begin{subfigure}{0.49\columnwidth}
\centering
\includegraphics[width=1\columnwidth]{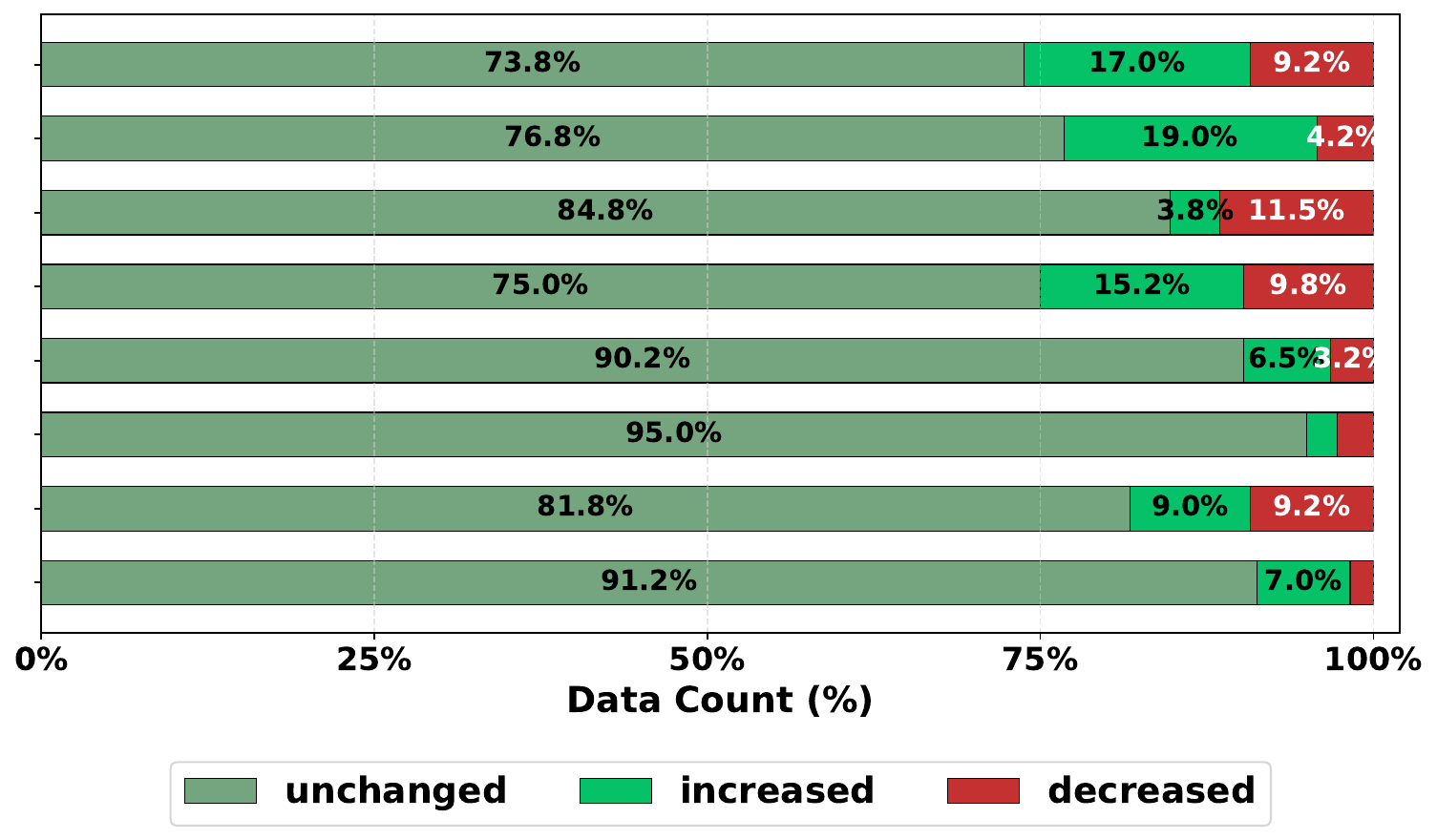}
\caption{Self Correction F1 Score Change.}
\label{fig:task2SChange}
\end{subfigure}
\caption{Self correction results of the different models for the \cons task.}
\label{fig:task2MultiTurn}
\end{figure*}

Finally, we report the self-correcting performance of the models for the \cons task after prompting them with,``Are you sure?''. To reduce the cost, we perform one run with all samples for all models. \autoref{fig:task2SC} shows that most models improve their performance upon re-evaluation, with \GeminiFlash achieving a notable gain of over 10\%. However, some models exhibit minimal change, such as \OOne, which shows only a 0.02\% reduction, while others experience a performance drop, like \LlamaThreeOneLarge and \GeminiPro, by 1.4\% and 3.8\%, respectively, the latter corresponding to a factor of $2.4x$.

\autoref{fig:task2SChange} provides a detailed breakdown of the changes in F1 scores for each model. The percentage here is calculated across all model outputs, regardless of their correctness. Most models exhibit no significant change in decision most of the time ($\geq 75\%$). Notably, \GeminiFlash shows the highest increase in F1 score at 19\% of the samples, while \GeminiPro demonstrates the most considerable decrease at 11.5\% of the samples.
\vspace{-2mm}
\subsection{\qa}

\begin{figure*}[!t]
\centering
\begin{subfigure}{0.49\columnwidth}
\centering
\includegraphics[width=1\columnwidth]{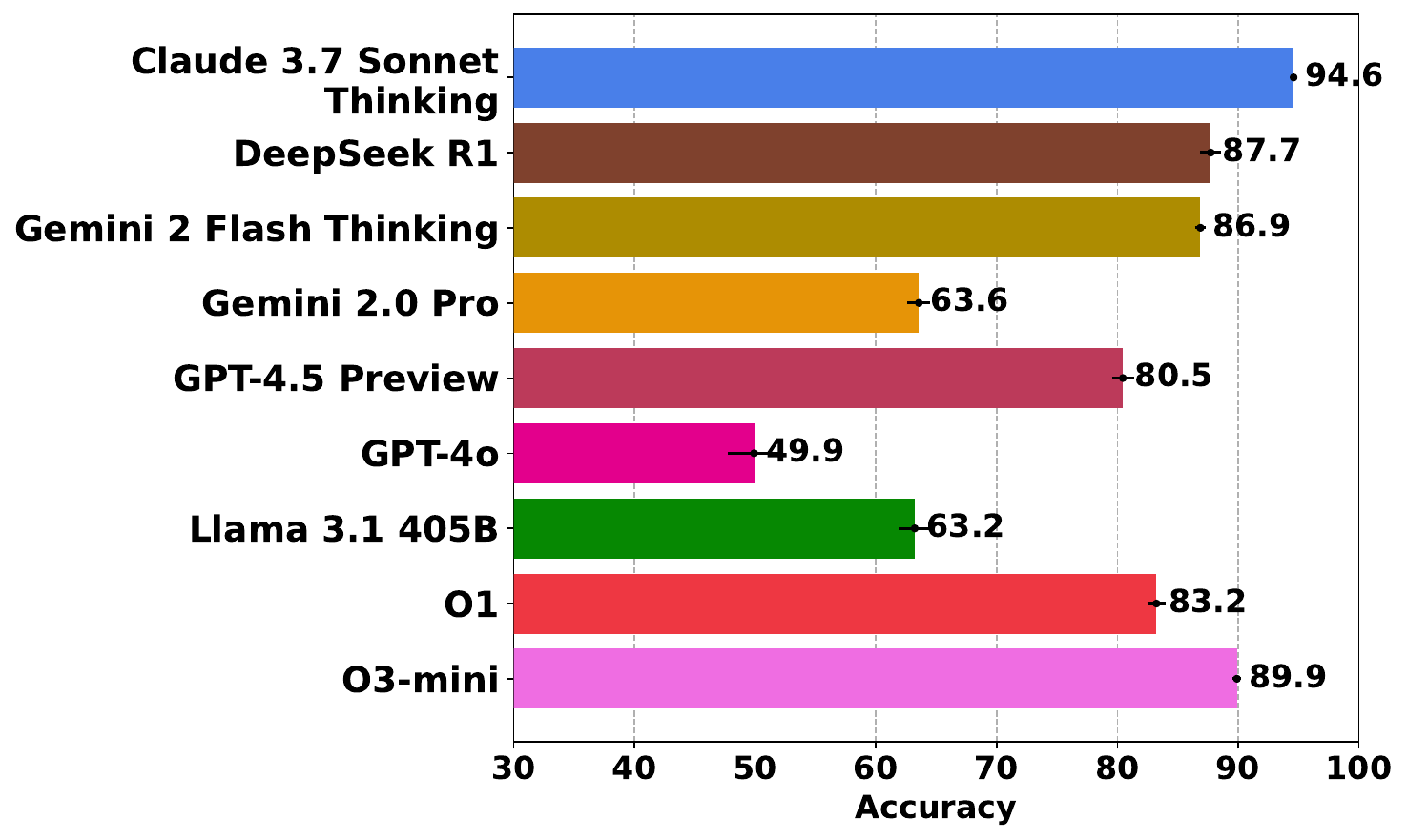}
\caption{Overall Accuracy.}
\label{fig:task3MainOverAll}
\end{subfigure}
\begin{subfigure}{0.50\columnwidth}
\centering
\includegraphics[width=1\columnwidth]{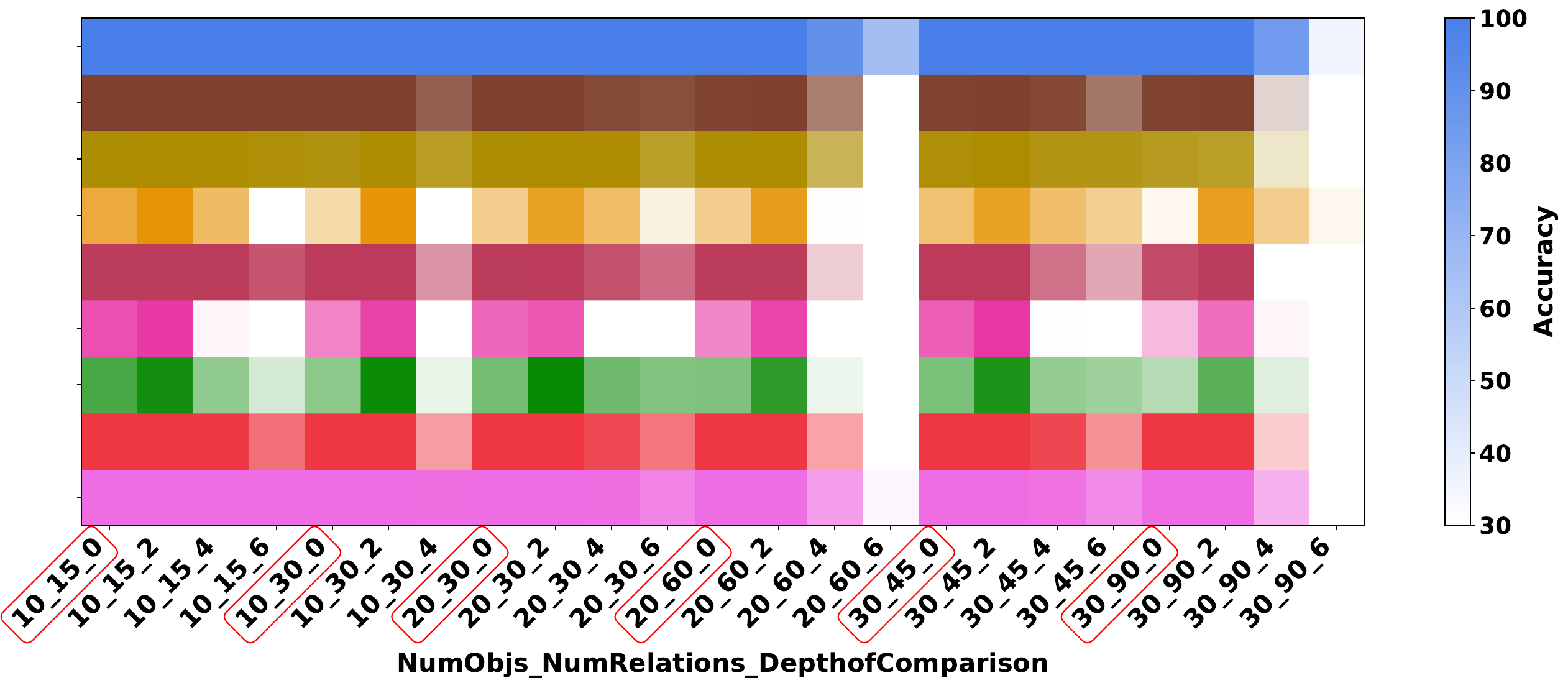}
\caption{Accuracy Per Group.}
\label{fig:task3MainFineGrained}
\end{subfigure}
\caption{Overview Accuracy of different models for the \qa task.}
\label{fig:task3Main}
\end{figure*}

Finally, we evaluate the \qa task. In this task, models are provided with a list of relations and a comparison statement, and are asked to determine whether the statement is ``True'', ``False'', or ``Unknown''. For this task, we only consider consistent graphs. During data generation, we ensure diversity in graph topology, object names, and the number of greater than ($>$) and smaller than ($<$) relations. We categorize the data into 23 different groups, each defined by varying numbers of objects, relations, and the minimum steps required to prove the relation in the question. In the group names, a depth of comparison of 0 indicates that the relation cannot be determined, hence the answer should be ``Unknown''. This setup results in a total of 460 samples. More details on the data groups and the prompt template can be found in \autoref{append:detailedEval}.

\autoref{fig:task3MainOverAll} displays the overall accuracy. Interestingly, for this task, the \OOne model drops from its second-place ranking in the previous tasks to below the \GeminiFlash model. Additionally, the top-performing models are much closer in performance for this task, with less than a 10\% gap between the first and fifth models. Notably, GPT 4.5 outperforms other instruction-based models, and indicates a smaller performance gap between reasoning and instruction-based models in this task. However, it is important to note that the baseline for random guessing is $\frac{1}{3}$; thus, while \GPTFourO achieves 49.9\% accuracy, this is only approximately 16.6\% better than random guessing.

Next, we present the fine-grained grouped results in \autoref{fig:task3MainFineGrained}. In this figure, the x-axis is labeled as $x\_y\_z$ represents groups covering the number of objects, the number of relations, and the depth of the comparison inference steps. Specifically we name the groups with ``Unknown'' comparison statement as 0 at $z$. Similar to the other tasks, as the number of objects and relations increases, the task becomes more challenging. It is also noteworthy that for instruction-based models such as \LlamaThreeOneLarge, \GPTFourO, and \GeminiPro, it is easier to determine the correct answer when it is not ``Unknown'', i.e., when the depth of comparison is greater than 0 in the chart.

\begin{figure*}[!t]
\centering
\begin{subfigure}{0.45\columnwidth}
\centering
\includegraphics[width=1\columnwidth]{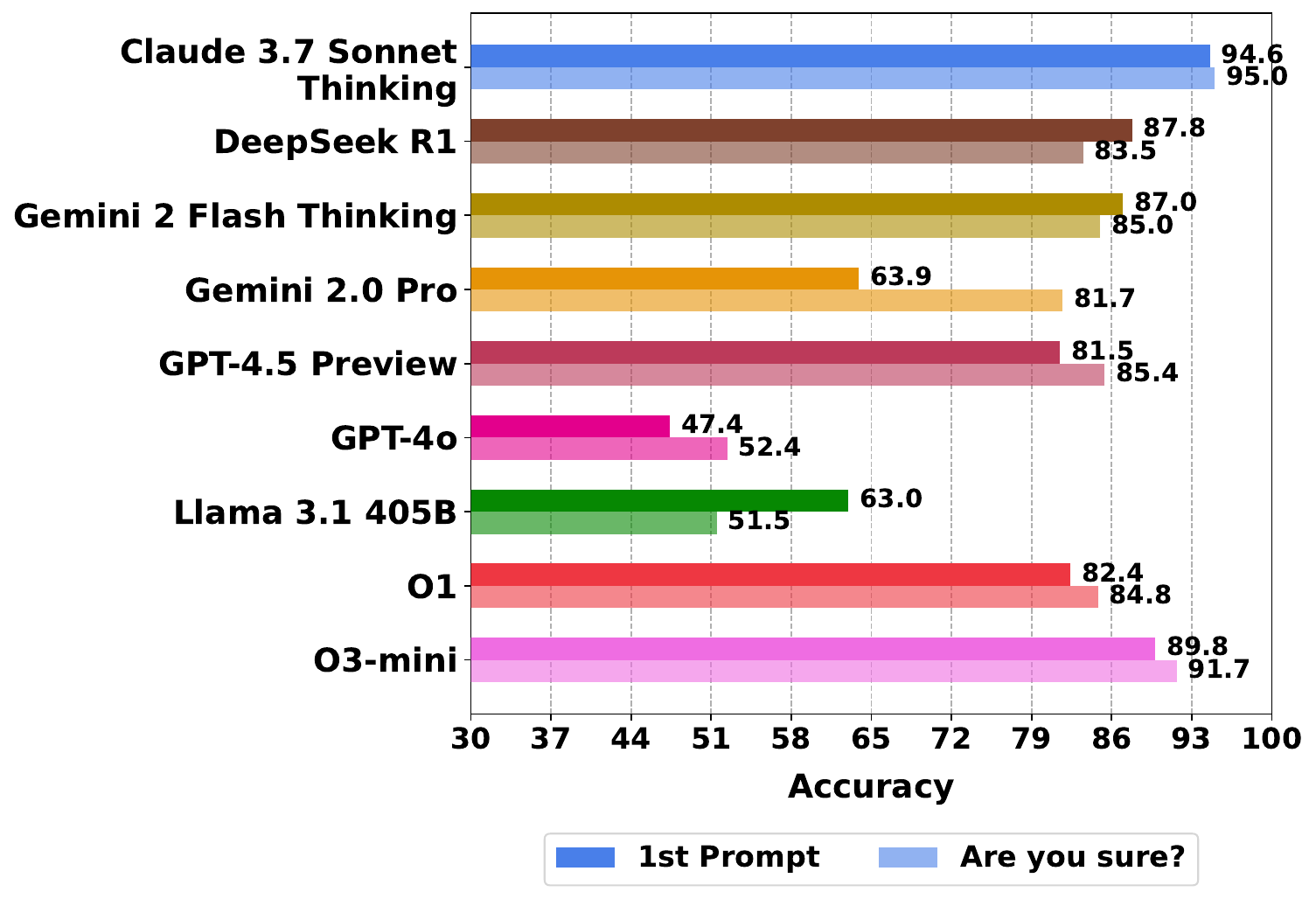}
\caption{Self Correction Accuracy.}
\label{fig:task3SC}
\end{subfigure}
\begin{subfigure}{0.54\columnwidth}
\centering
\includegraphics[width=1\columnwidth]{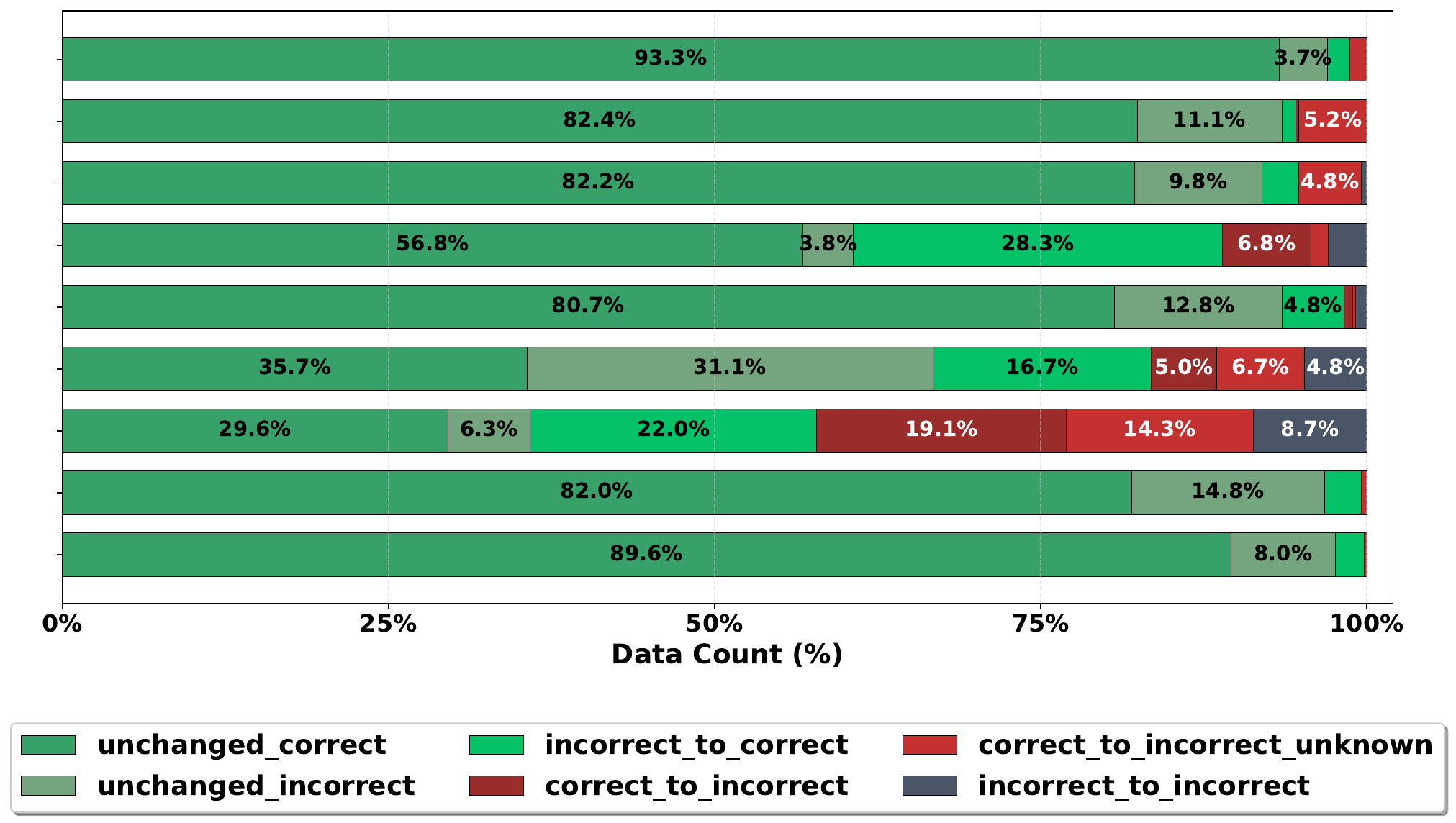}
\caption{Self Correction Change in Detail.}
\label{fig:task3SChange}
\end{subfigure}
\caption{Self correction results of the different models for the \qa task.}
\label{fig:task3MultiTurn}
\end{figure*}

We also plot the models' self-correction results in \autoref{fig:task3SC}. To reduce the cost, we perform one run with all samples for all models. \autoref{fig:task3SC} displays both the original results and those obtained after asking, ``Are you sure?''. Surprisingly, for this task, \GeminiPro, which performed the worst in self-improvement during the \cons task, shows the best improvement, with a gain of 17.8\%, while \LlamaThreeOneLarge continues to perform poorly after the self-check, with a drop of 11.5\%. Other models slightly improved, except for \ROne and \GeminiFlash, which showed slight degradations.

\autoref{fig:task3SChange} illustrates the fine-grained changes for all models. Additionally, we highlight instances where the model predicts ``Unknown'' and the correct label is \textbf{not} ``Unknown''. As the figure shows, the instruction-based models exhibit the most incorrect changes to ``Unknown'', with \LlamaThreeOneLarge and \GPTFourO incorrectly converting 14.3\% and 6.7\% of the correct decisions to ``Unknown'', respectively.

\vspace{-2mm}

\vspace{-2mm}
\section{Related Work}
\vspace{-2mm}
\label{sec:related_work}
\paragraph{Logical Reasoning in LLMs.} Logical reasoning has been studied as a fundamental capability of large language models in the context of inductive reasoning, propositional and first-order logic~\citep{patel2024multi,ryu2024divide,parmar2024logicbench,xu2024faithful}, and a diverse set of logical puzzles~\citep{lin2025zebralogic,shah2024causal}. Similarly to \benchmark, these works often involve one or multi-hop reasoning as a way of testing whether models can retrieve and reason upon more than one statement or fact presented in context~\citep{trivedi2022musique,yang2018hotpotqa,patel2024multi,schnitzler2024morehopqa}. For example, HotpotQA includes a set of linguistic facts in each example, and then asks related factoid comparison questions. However, recent work has also shown that the accuracy of models in conducting multi-hop reasoning drops as the length of context increases~\citep{levy2024same,balachandran2024eureka} or with the number of inference rules~\citep{patel2024multi}.

Relational reasoning in LLMs has not been studied extensively yet since until recently the task was beyond the capabilities of the state-of-the-art, especially for larger problem sizes. Preliminary work~\citep{li2024llms,alotaibi2024graph} with a few-hop queries has shown that larger models can conduct simple logical queries on graphs to extract relationships between objects or their attributes. Advancements in test-time compute and post-training (or self-training) via Reinforcement Learning (RL)~\citep{lightman2023let,wang2023math,gulcehre2023reinforced} have shown to significantly improve general reasoning skills for problems like math and coding~\citep{guo2025deepseek,O3mini,jaech2024openai}. Qualitatively, these methods lengthen the generations of models in the form of extended reasoning traces, which contain notions of self-reflection, backtracking, and exploration. In this work, we include several of the state-of-the-art reasoning models, to better understand whether these techniques generalize to more complex forms of reasoning, such as inference and cycle detection in larger graphs. More relatedly to our work, \cite{abbe2024far} has shown that intermediate steps in detailed scratchpads in transformers can help address barriers introduced by high connectivity and longer chains in difficult queries.

\paragraph{Planning in LLMs.} Our work can also be situated in the context of studying planning abilities of LLMs, which has been evaluated extensively in previous work~\citep{valmeekam2024llms,valmeekam2023planning,wang2024planning,zheng2024natural} with benchmarks like PlanBench~\citep{valmeekam2024llms}, NaturalPlan~\citep{zheng2024natural}, CogEval~\citep{momennejad2023evaluating}. CogEval in particular also looks at routing problems in graphs, but does not extend to cycle detection or consistency checks of relationships. \cite{wang2024pictureworththousandwords} instead formulates a task in a multimodal setting that asks multiple-choice questions about the relationships between pairs of objects, which can be inferred from other relationships already presented in context.

Most importantly, given the high computational complexity of the problems in \benchmark, our benchmark also contributes to an emerging line of work that studies how transformers solve NP-hard and NP-complete problems~\citep{fan2023nphardeval}. With the increased generality of current architectures, it is important to understand in detail the limitations of transformers in solving such complex problems and how they scale with the difficulty of the problem.%

\vspace{-2mm}
\section{Conclusion}
\vspace{-2mm}
In conclusion, our work demonstrates that while current LLMs have made notable strides in handling structured relational reasoning, significant performance gaps remain—particularly as task complexity increases. \benchmark not only reveals the limitations of instruction-based models in generating logically consistent plans but also highlights the robust, albeit varied, performance of dedicated reasoning models. This benchmark serves as both a diagnostic tool and a catalyst for future research, guiding the development of more sophisticated architectures and self-correction strategies that can better tackle real-world, complex logical planning problems.

\newpage

\section*{Reproducibility Statement}
\paragraph{Implementation.} All our experimental pipelines are implemented using an open-source evaluation framework. We will make all data generation scripts publicly available upon publication, which will also enable sampling other versions \benchmark with different problem sizes. The open source framework we use supports parameterized scripts that not only facilitate reproducibility but also allow for the generation of more challenging and customized setups tailored to specific scenarios, such as particular object names and varying plan lengths. Additionally, the inference configurations used for all tasks are presented in \autoref{tab:models}, and the prompts used for each task are detailed in \autoref{append:detailedEval}. We experiment with different rephrasing of the prompts but found no significant effect on the outcomes. For all experiments we run, Top\_p is set to 0.95, presence\_penalty is set to 0 by default if the model supports this.

\paragraph{Note on experimental results.} Our experiments on \GeminiPro for \qa were interrupted after four runs as the model was softly deprecated upon the release of Gemini 2.5 Pro on March 25, 2025. Tasks \gen and \cons instead include results for five runs for \GeminiPro. Additionally, due to restrictive rate limiting imposed by Claude, we were unable to complete all planned runs with Claude models before this submission. The result presented in \qa is only based on one run on Claude 3.7 Sonnet with thinking model enabled.

\begin{table}[t]
  \centering
  \footnotesize
  \caption{List of models studied in this study and corresponding temperature and maximum token limits used for all experiments.}
    \begin{tabular}{lcrc}
    \toprule
    \bfseries{Model} & \bfseries{temp.} & \bfseries{max token} & \bfseries{reasoning} \\
    \hline
    Claude 3.7 Sonnet Thinking 2025-02-19 \citep{Claude37Sonnet} & 1.0     & 32,768 & y \\
    \hline
    \ROne \citep{guo2025deepseek} & 0.6   & 32,768 & y \\
    \hline
    \GeminiPro Exp 2025-02-05 \citep{Gemini2Pro} & 1.0     & 4,096  & n \\
    \hline
    \GeminiFlash Exp 2025-01-21 \citep{GeminiFlash}& 1.0     & 32,768 & y \\
    \hline
    \OOne  2024-12-17  \citep{jaech2024openai}& NA    & 32,000   & y \\
    \hline
    \OThree 2025-01-31 (high) \citep{O3mini} & NA    & 32,000   & y \\
    \hline
    \GPTFourO 2024-08-06 \citep{hurst2024gpt} & 1.0     & 4,096  & n \\
    \hline
    \GPTFourFive Preview 2025-02-27 \citep{hurst2024gpt} & 1.0     & 4,096  & n \\
    \hline
    \LlamaThreeOneLarge \citep{dubey2024llama} & 1.0     & 4,096  & n \\
    \bottomrule
    \end{tabular}%
  \label{tab:models}%
\end{table}%

\bibliography{colm2025_conference}
\bibliographystyle{colm2025_conference}

\appendix
\clearpage
\section*{Appendix}

\section{Detailed Evaluation Settings}
\label{append:detailedEval}
We first present a detailed list of various object and relation numbers, together with other settings for each task.

\subsection{\gen}
\label{append:detailedEvaltask1}
The combination of the number of objects and the number of relations for \gen task:
\begin{promptbox}[\gen data groups: \\ NumObjects\_NumRelations]
3_3, 
5_5, 5_7, 5_9, 
10_10, 10_15, 10_20, 10_25, 10_30, 10_35, 10_40, 10_45, 
15_15, 15_24, 15_34, 15_43, 15_53, 15_62, 15_72, 15_81, 15_91, 15_100,
20_20, 20_29, 20_38, 20_47, 20_56, 20_65, 20_74, 
20_83, 20_92, 20_101, 20_109, 20_118, 20_127, 20_136, 
20_145, 20_154, 20_163, 20_172, 20_181, 20_190, 
25_25, 25_34, 25_43, 25_53, 25_62, 25_71, 25_80, 25_89, 
25_99, 25_108, 25_117, 25_126, 25_136, 25_145, 25_154, 
25_163, 25_172, 25_182, 25_191, 25_200, 
30_30, 30_39, 30_48, 30_57, 30_66, 30_75, 30_84, 30_93,
30_102, 30_111, 30_119, 30_128, 30_137, 30_146, 30_155,
30_164, 30_173, 30_182, 30_191, 30_200, 
40_40, 40_49, 40_58, 40_67, 40_76, 40_85, 40_94, 40_103,
40_112, 40_121, 40_130, 40_139, 40_148, 40_157, 40_166,
40_174, 40_183, 40_192, 40_201, 40_210, 40_219, 40_228, 
40_237, 40_246, 40_255, 40_264, 40_273, 40_282, 40_291, 40_300, 
50_50, 50_59, 50_67, 50_76, 50_84, 50_93, 50_102, 50_110,
50_119, 50_128, 50_136, 50_145, 50_153, 50_162, 50_171, 
50_179, 50_188, 50_197, 50_205, 50_214, 50_222, 50_231, 
50_240, 50_248, 50_257, 50_266, 50_274, 50_283, 50_291, 50_300
\end{promptbox}

The prompt template we use with these groups in task \gen:
\begin{promptbox}[\gen task prompt template]
You are tasked with generating a list of logical and consistent 
relationships between objects:
- Each object should have a unique name. 
- The relationships use either the "<" (smaller than) 
  or ">" (larger than) operator.
- The relationships must be logically consistent and acyclic 
  (i.e., no cycles like A > B > A).
- Not all objects need to be directly related, but all 
  relationships must be valid.
   
Now generate a logical and consistent list of {{num_relations}} 
relationships with {{num_objects}} objects.
   
Output Format:
Return only a valid JSON object after the <RESPONSE> tag, 
structured as follows:
<RESPONSE>
{
   "relationships": [
       "<object> <operator> <object>",
       "<object> <operator> <object>",
        ...
       "<object> <operator> <object>"
   ]
}
</RESPONSE> 
\end{promptbox}

\subsection{\cons}
\label{append:detailedEvaltask2}
We provide the groups of the number of objects, the number of relations, and the minium cycle length of the directed graph. We generate 20 unique graphs for each group.

\begin{promptbox}[\cons data groups: \\ NumObjects\_NumRelations\_MinCycleLength]
10_15_0, 10_15_3, 10_15_6,
10_30_0, 10_30_3, 10_30_6,
20_30_0, 20_30_6, 20_30_12, 
20_60_0, 20_60_6, 20_60_12, 
30_45_0, 30_45_6, 30_45_12, 30_45_24, 
30_90_0, 30_90_6, 30_90_12, 30_90_24
\end{promptbox}

The prompt template we use in task \cons:
\begin{promptbox}[\cons task prompt template]
Here is a list of relationships {{relationships}}, your job is to decide
if there are any contradictions or cycles in them, if yes (meaning there
are contradictions in the list), list all cycles that cause the
contradiction after the yes or no answer. Don't write code to solve
this problem. You can do whatever analysis that you need at the
beginning, but make sure to always finish your response with the
following format example, after starting with the OUTPUT tag, don't add
any additional text or comments.
OUTPUT:
Yes or No (Yes means there are contradictions, No means there is not)
1. Cycle: <A, B, C, D, E, A> 
2. Cycle: <cycle>
...
N. Cycle: <cycle>
\end{promptbox}

\begin{promptbox}[\cons task self-correction prompt template]
Are you sure? You can do whatever analysis that you need at the 
beginning, but make sure to always finish your response with the 
following format example, after starting with the OUTPUT tag, 
don't add any additional text or comments.
OUTPUT:
Yes or No (Yes means there are contradictions, No means there is not)
1. Cycle: <A, B, C, D, E, A> 
2. Cycle: <cycle>
...
N. Cycle: <cycle>
\end{promptbox}

\begin{figure}[!t]
    \centering
    \includegraphics[width=0.3\textwidth]{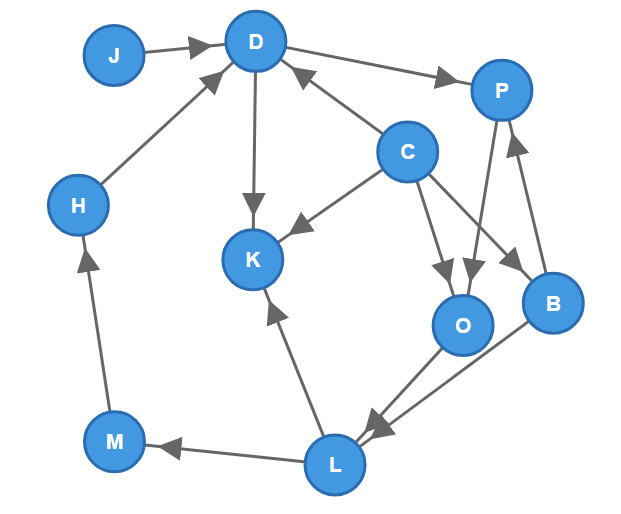}
    \caption{An example graph for the \cons task.}
    \label{fig:graph-example}
\end{figure}

As an example, \autoref{fig:graph-example} illustrates an example of the graph used in the benchmark dataset. The relations represented in the prompt are $C > D, O > L, M > H, M < L, D > P, C > O, D < J, D < H, C > K, B > L, L > K, K < D, B > P, O < P, C > B$. This graph contains a single cycle with 6 nodes: $M > H > D > P > O > L > M$. 

\subsection{\qa} 
\label{append:detailedEvaltask3}
We provide the groups of the number of objects, the number of relations, and the steps between two objects in the comparison statement.

\begin{promptbox}[\qa data groups: \\ NumObjects\_NumRelations\_DepthofComparison]
10_15_0, 10_15_2, 10_15_4, 10_15_6, 
10_30_0, 10_30_2, 10_30_4, 
20_30_0, 20_30_2, 20_30_4, 20_30_6, 
20_60_0, 20_60_2, 20_60_4, 20_60_6, 
30_45_0, 30_45_2, 30_45_4, 30_45_6, 
30_90_0, 30_90_2, 30_90_4, 30_90_6
\end{promptbox}

The prompt template we use in the task \qa:
\begin{promptbox}[\qa task prompt template]
Given a set of relationship statements:
{{relationships}}
   
And a comparison statement to evaluate:
{{comparison}}
   
Your task is to determine if the comparison is True, False, 
or Unknown based solely on the provided relationships.
    
Guidelines for evaluation:
- True: The comparison can be logically deduced from the given 
  relationships
- False: The comparison can be proven incorrect using the given
  relationships
- Unknown: There is insufficient information in the relationships
  to definitively prove the comparison true or false
    
Important notes:
- Use ONLY the information provided in the relationships
- Do not make assumptions beyond what is explicitly stated
- If there are multiple possible interpretations, mark as Unknown
- You can do whatever analysis that you need in the response, 
  but your response must end with the OUTPUT format below
- Do not add any additional text after the output
    
Required output format:
OUTPUT:
True/False/Unknown
\end{promptbox}

\begin{promptbox}[\qa task self-correction prompt template]
Are you sure? 
Note: You can do whatever analysis that you need in the response, 
but your response must end with the OUTPUT format below. 
Do not add any additional text after the output.
Required output format:
OUTPUT:
True/False/Unknown
\end{promptbox}

\section{Detailed Task Results}
\label{append:detailedEvalResult}

\subsection{Detailed \gen Task Results}
\label{append:detailedEvalResulttask1}

\begin{sidewaysfigure}[!t]
\centering
\begin{subfigure}{1\columnwidth}
\centering
\includegraphics[width=1\columnwidth]{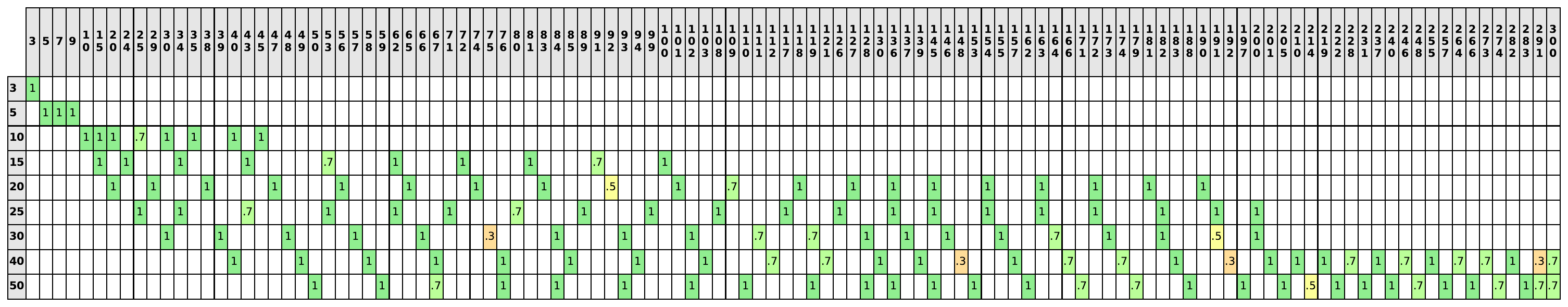}
\caption{Deepseek R1.}
\label{fig:task1deepseek}
\end{subfigure}
\begin{subfigure}{1\columnwidth}
\centering
\includegraphics[width=1\columnwidth]{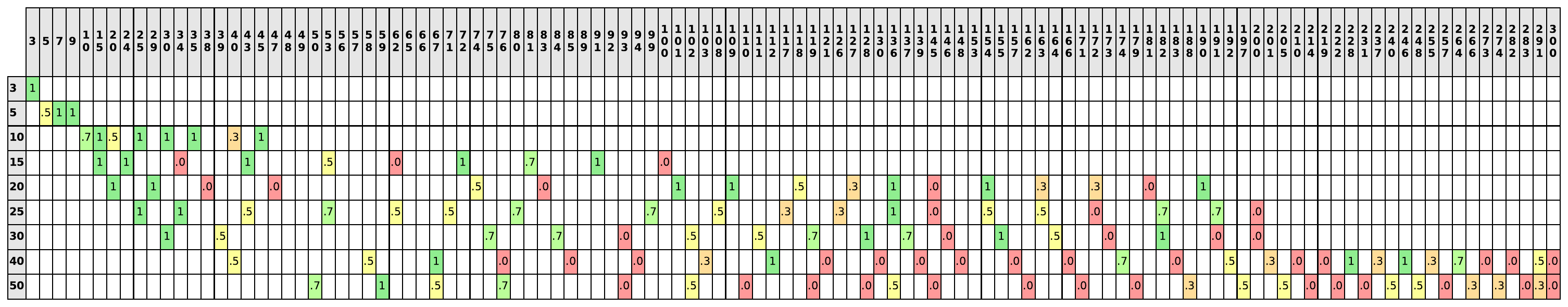}
\caption{Gemini 2 Flash Thinking.}
\label{fig:task1genimithinking}
\end{subfigure}
\caption{Detailed average accuracy of \gen task}
\label{fig:task1detail}
\end{sidewaysfigure}

\begin{sidewaysfigure}[!t]
\centering
\begin{subfigure}{1\columnwidth}
\centering
\includegraphics[width=1\columnwidth]{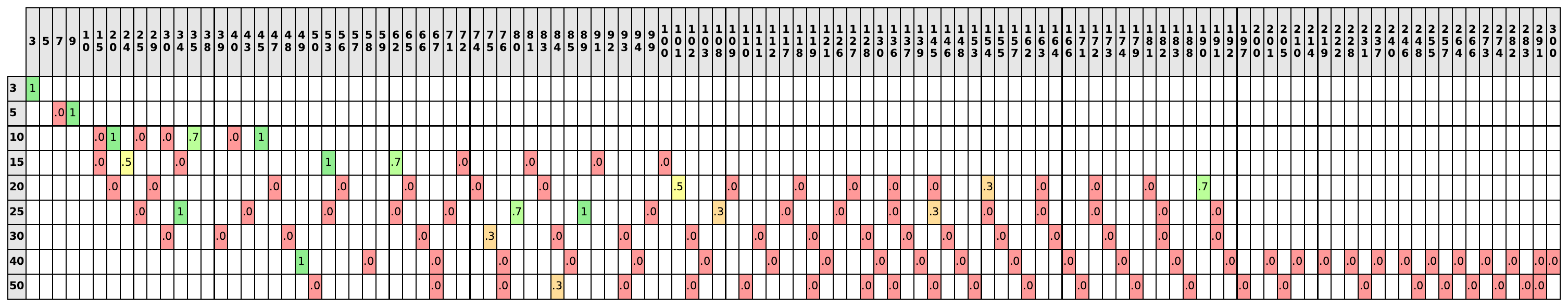}
\caption{Gemini 2.0 Pro.}
\label{fig:task1genimipro}
\end{subfigure}
\begin{subfigure}{1\columnwidth}
\centering
\includegraphics[width=1\columnwidth]{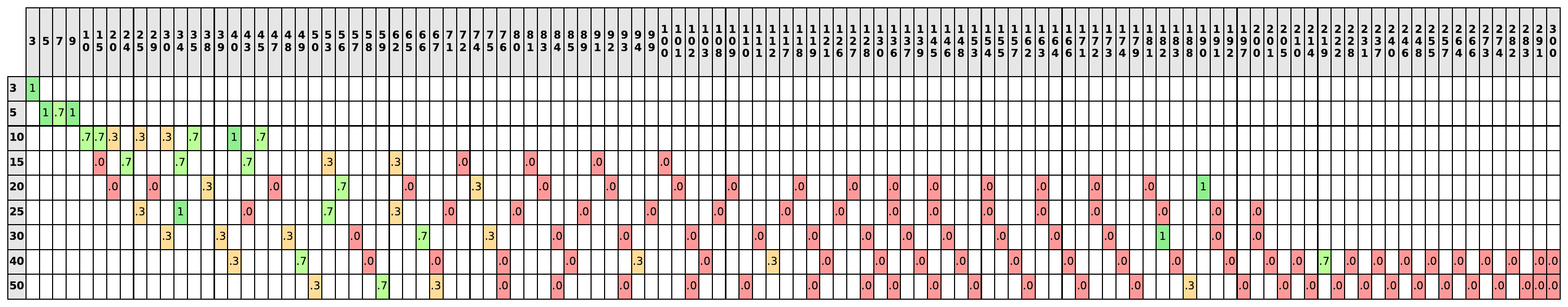}
\caption{GPT-4.5 Preview.}
\label{fig:task1gpt45}
\end{subfigure}
\caption{Detailed average accuracy of \gen task - Continued}
\label{fig:task1detail1}
\end{sidewaysfigure}

\begin{sidewaysfigure}
\centering
\begin{subfigure}{1\columnwidth}
\centering
\includegraphics[width=1\columnwidth]{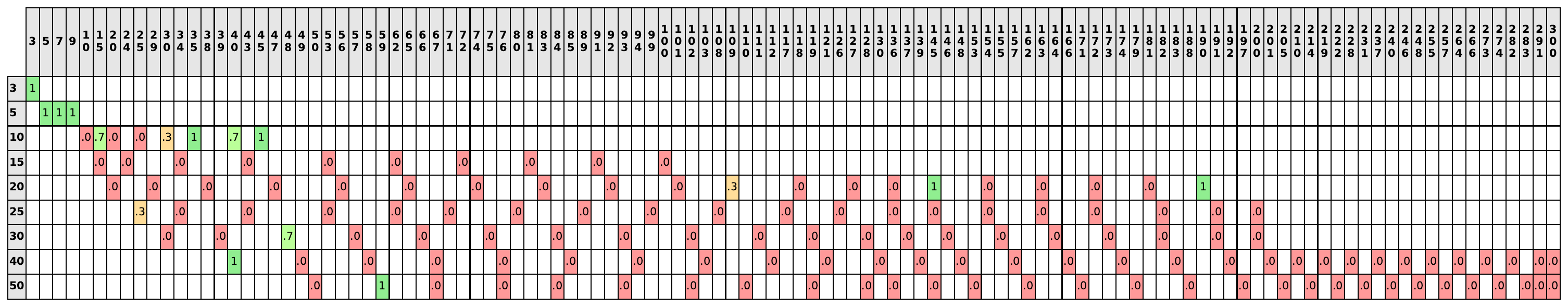}
\caption{GPT-4o.}
\label{fig:task1gpt4o}
\end{subfigure}
\begin{subfigure}{1\columnwidth}
\centering
\includegraphics[width=1\columnwidth]{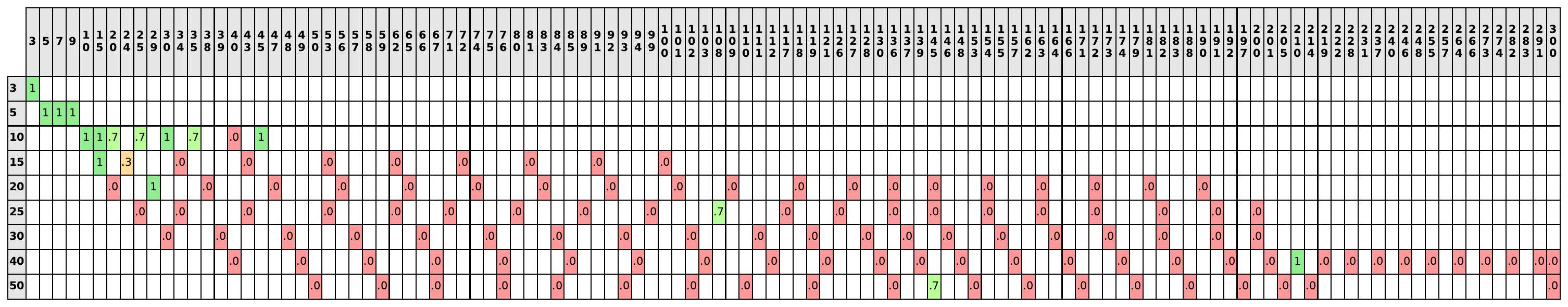}
\caption{Llama 3.1 405B Instruct.}
\label{fig:task1llama}
\end{subfigure}
\caption{Detailed average accuracy of \gen task - Continued}
\label{fig:task1detail2}
\end{sidewaysfigure}

\begin{sidewaysfigure*}[!t]
\begin{subfigure}{1\columnwidth}
\centering
\includegraphics[width=1\columnwidth]{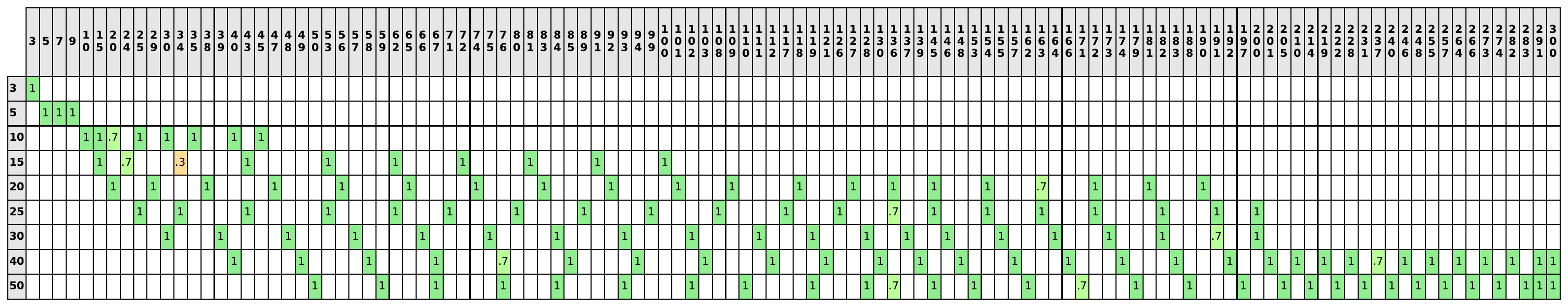}
\caption{O1.}
\label{fig:task1o1}
\end{subfigure}
\begin{subfigure}{1\columnwidth}
\centering
\includegraphics[width=1\columnwidth]{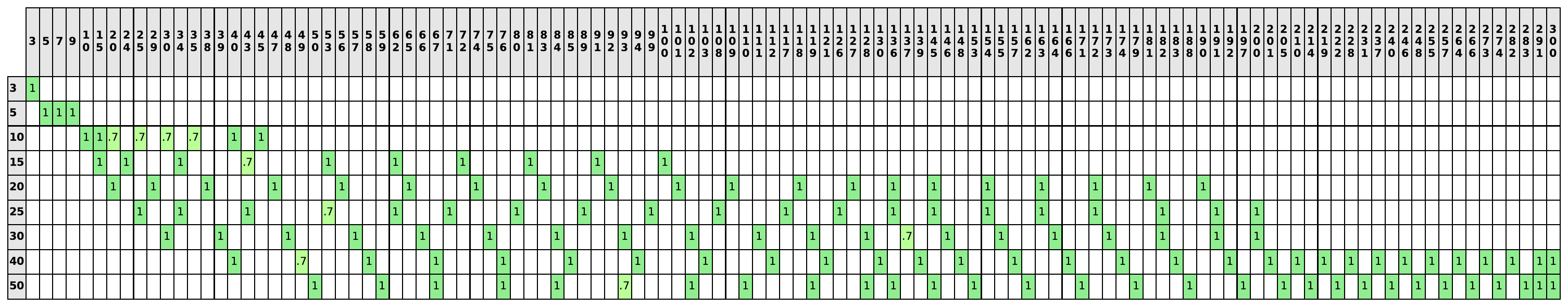}
\caption{O3-MINI.}
\label{fig:task1o3}
\end{subfigure}
\caption{Detailed Average Accuracy of \gen Task - Continued}
\label{fig:task1detail3}
\end{sidewaysfigure*}

\autoref{fig:task1detail}, \autoref{fig:task1detail1}, \autoref{fig:task1detail2} and \autoref{fig:task1detail3} show the detailed result of \gen task. The rows are the number of objects, and the columns are the number of relations prompted to the model. The empty cells are the combinations that are not covered in this test. Each colored cell shows the average accuracy result of 3 runs.

\begin{figure}
\centering
\includegraphics[width=1\columnwidth]{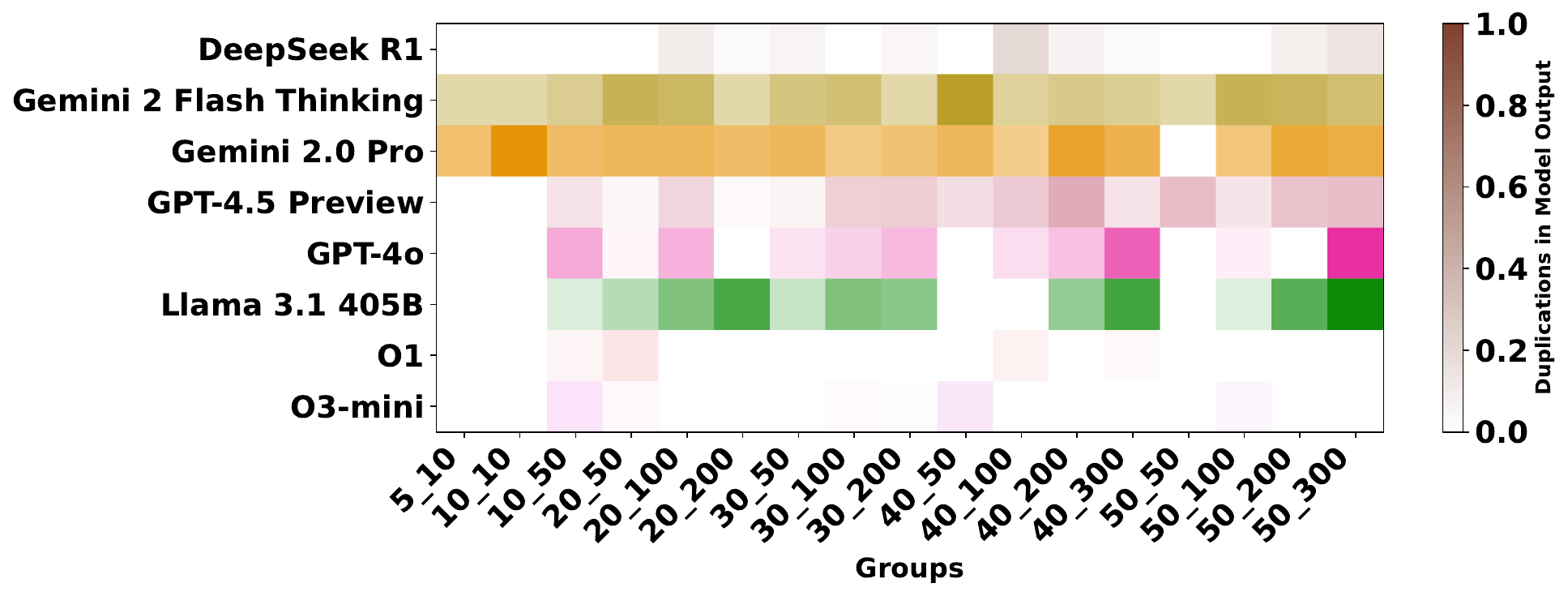}
\caption{\gen Relation Duplications in Model Output}
\label{fig:task1dup}
\end{figure}

We also notice that some models are easier to output duplicated relations in general, like Gemini models. And some models get easier to output duplicate relations as the task gets harder, such as \GPTFourO and \LlamaThreeOneLarge. \autoref{fig:task1dup} shows the detailed results of the average existence of duplications (True of False) in model's output for each model and task groups.

\subsection{Detailed \cons Task Result}
\label{append:detailedEvalResulttask2}

\begin{figure*}[!t]
\centering
\begin{subfigure}{0.49\columnwidth}
\centering
\includegraphics[width=1\columnwidth]{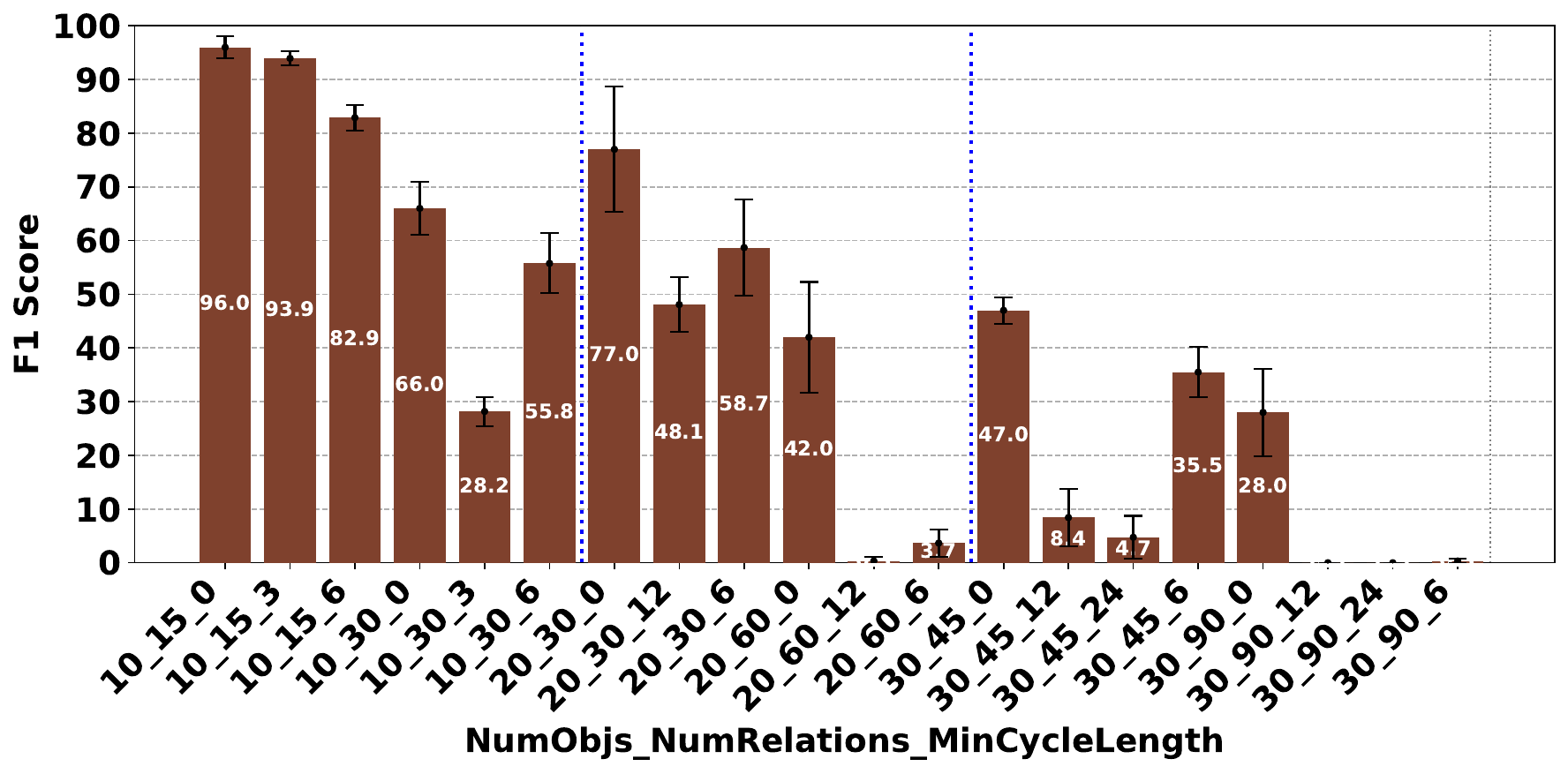}
\caption{Deepseek R1.}
\label{fig:task2deepseek}
\end{subfigure}
\begin{subfigure}{0.49\columnwidth}
\centering
\includegraphics[width=1\columnwidth]{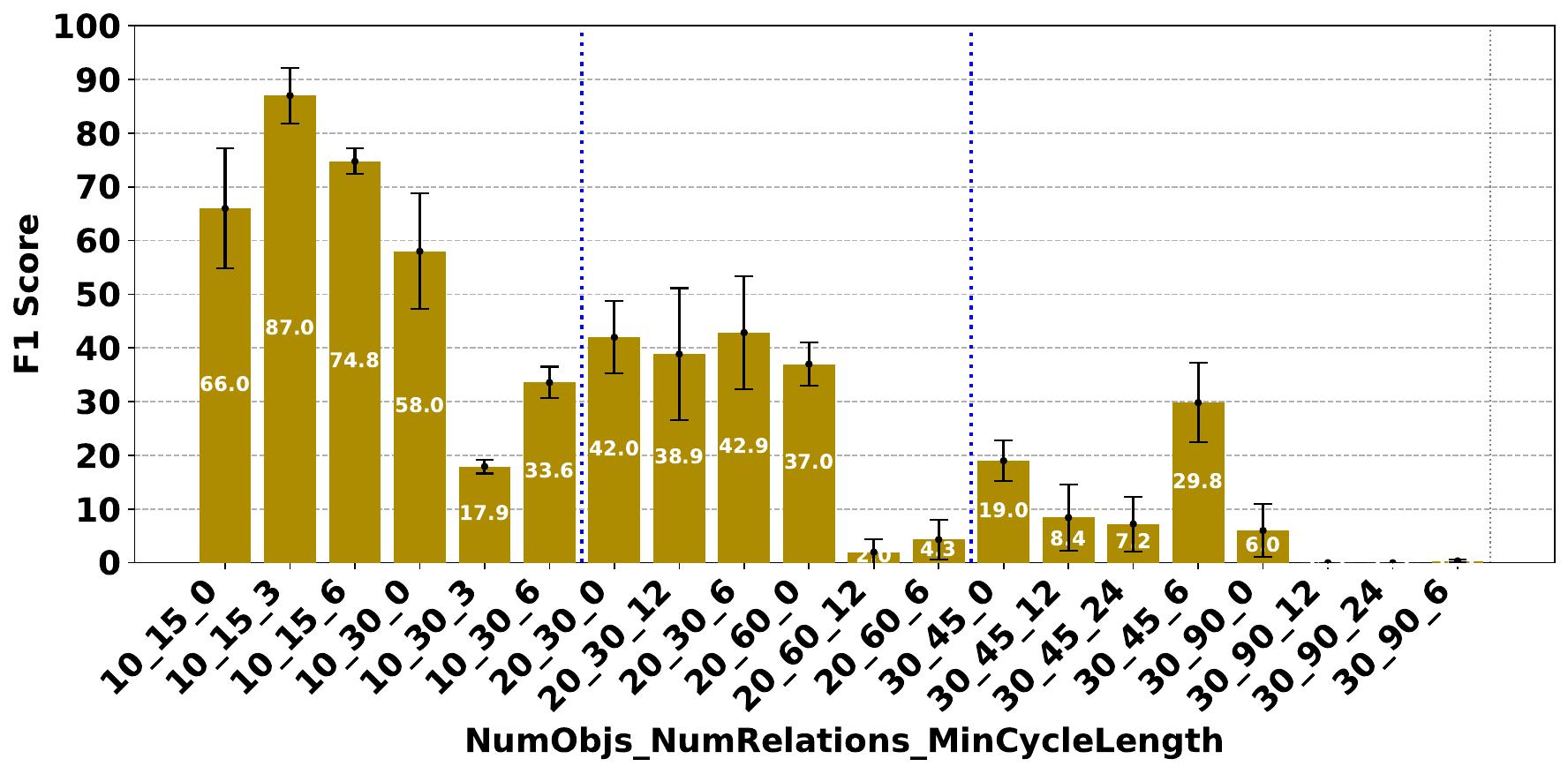}
\caption{Gemini 2 Flash Thinking.}
\label{fig:task2genimithinking}
\end{subfigure}
\begin{subfigure}{0.49\columnwidth}
\centering
\includegraphics[width=1\columnwidth]{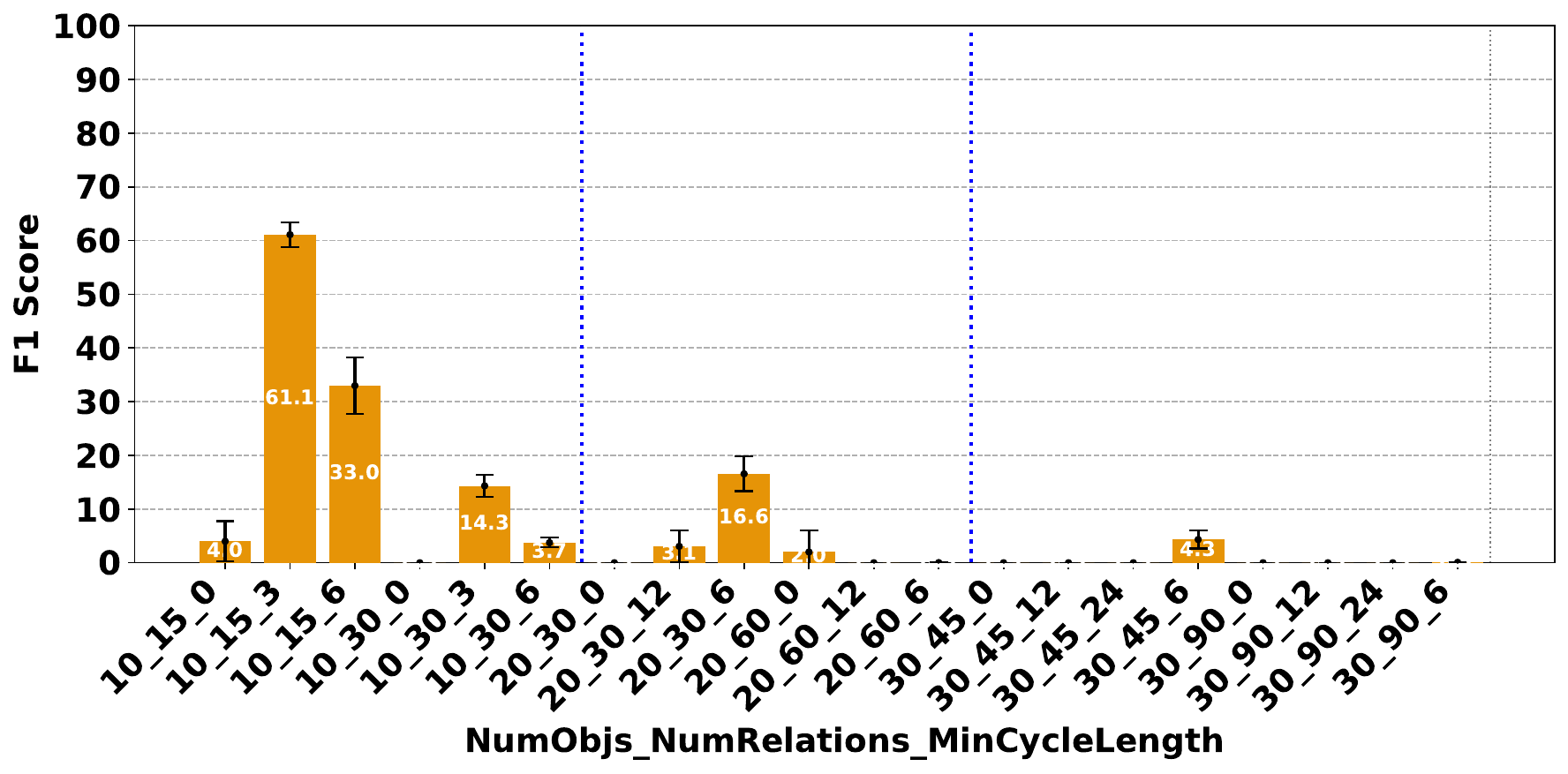}
\caption{Gemini 2.0 Pro.}
\label{fig:task2genimipro}
\end{subfigure}
\begin{subfigure}{0.49\columnwidth}
\centering
\includegraphics[width=1\columnwidth]{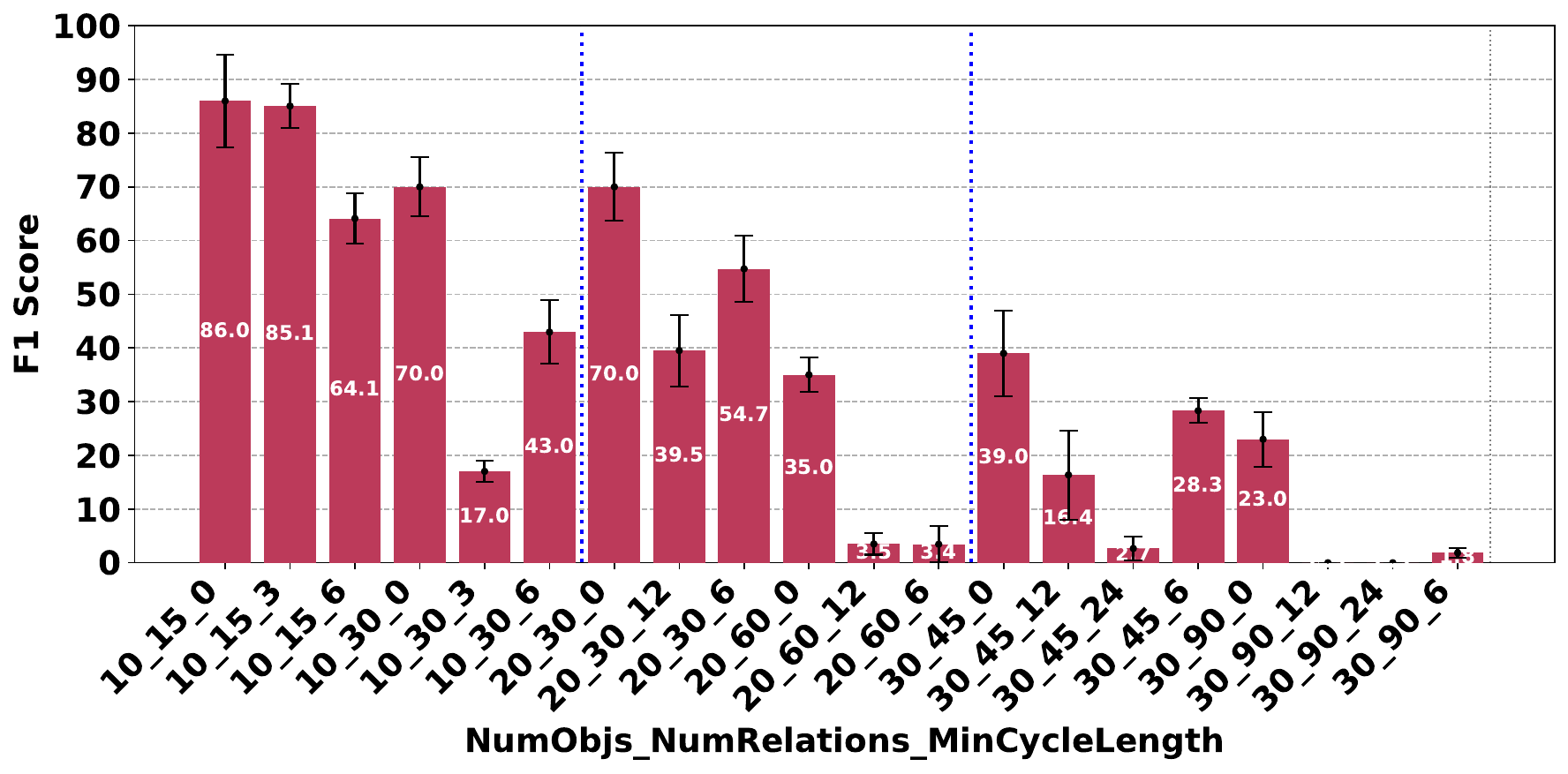}
\caption{GPT-4.5 Preview.}
\label{fig:task2gpt45}
\end{subfigure}
\centering
\begin{subfigure}{0.49\columnwidth}
\centering
\includegraphics[width=1\columnwidth]{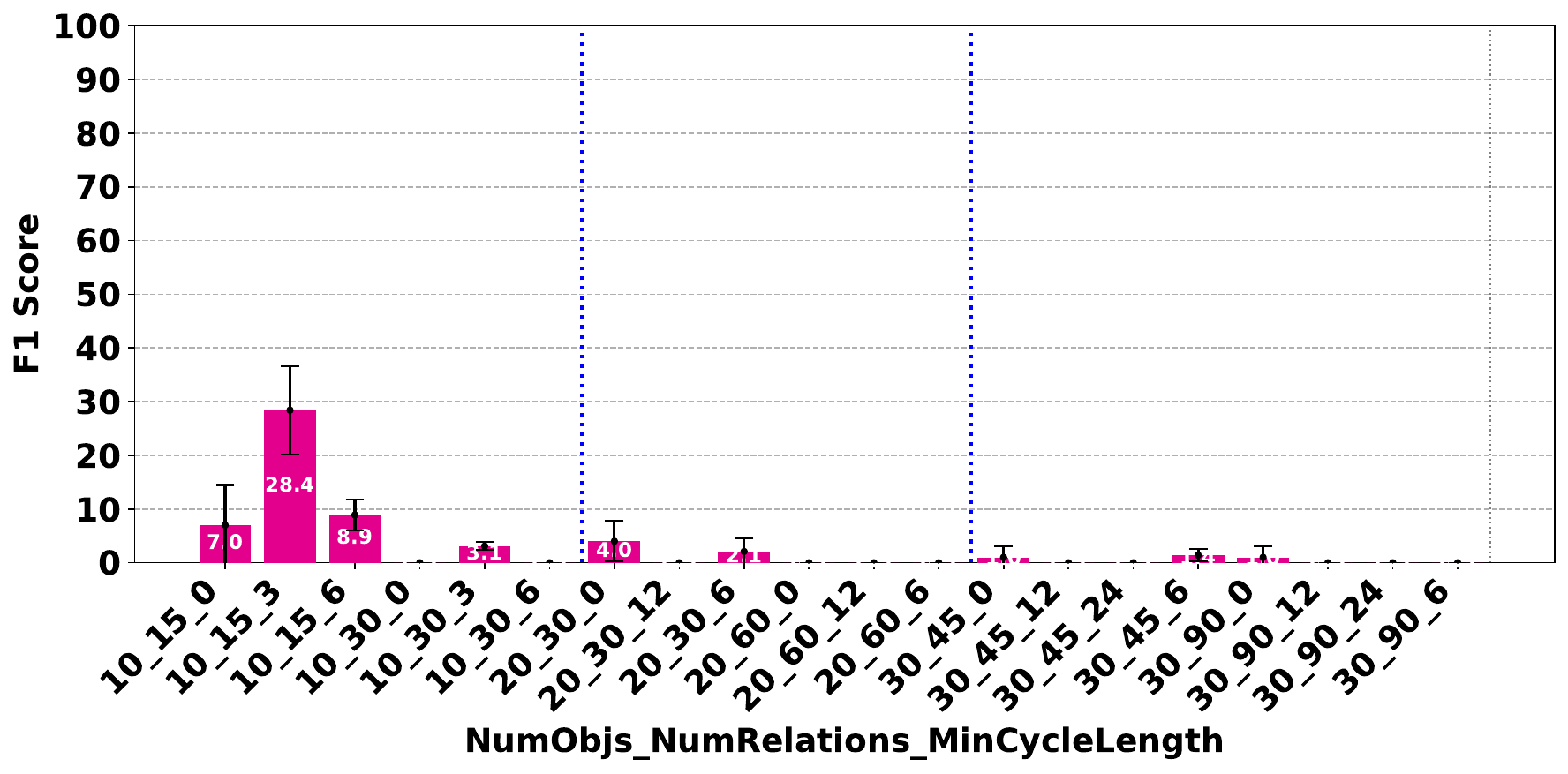}
\caption{GPT-4o.}
\label{fig:task2gpt4o}
\end{subfigure}
\begin{subfigure}{0.49\columnwidth}
\centering
\includegraphics[width=1\columnwidth]{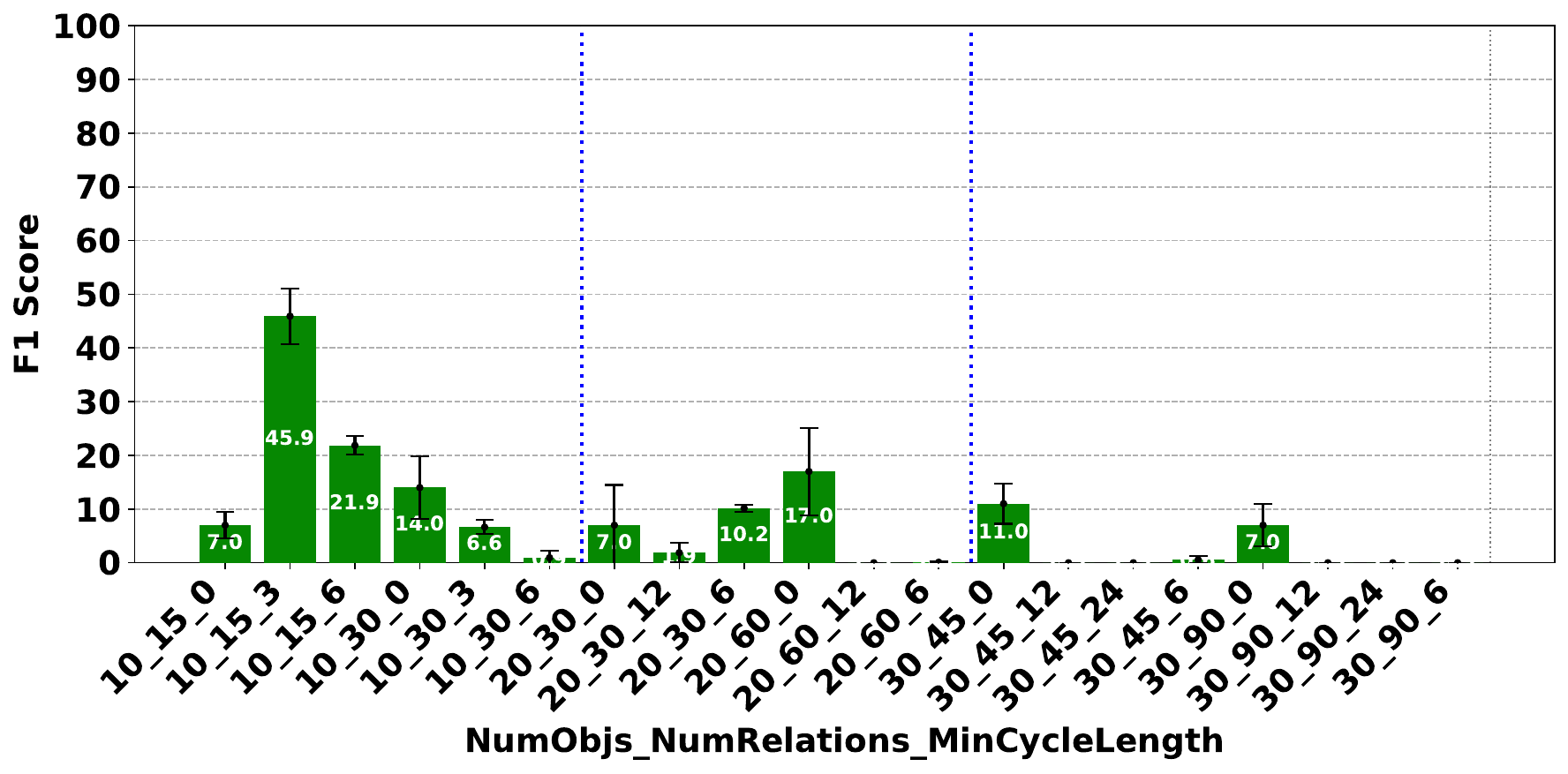}
\caption{Llama 3.1 405B Instruct.}
\label{fig:task2llama}
\end{subfigure}
\begin{subfigure}{0.49\columnwidth}
\centering
\includegraphics[width=1\columnwidth]{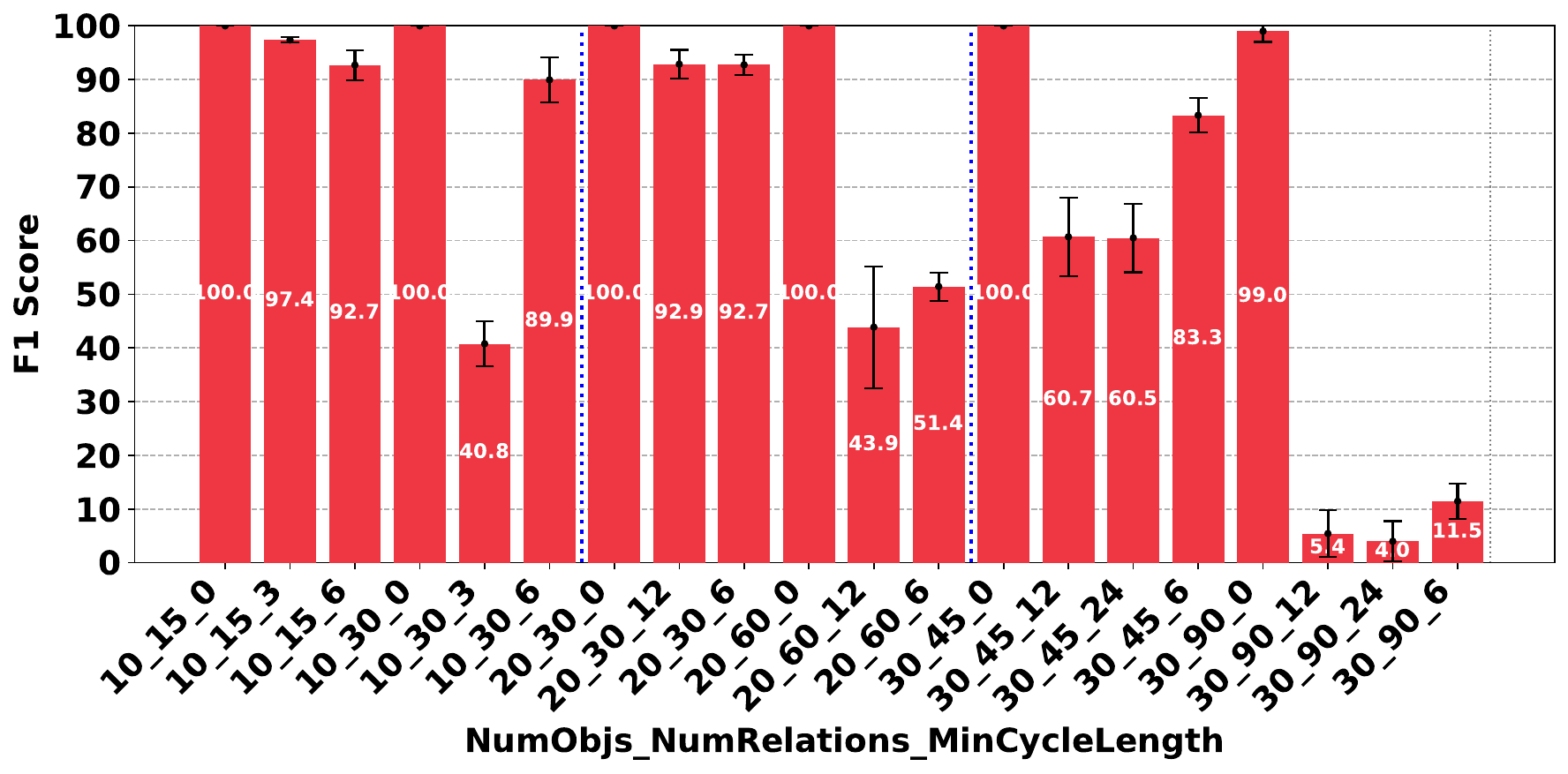}
\caption{O1.}
\label{fig:task2o1}
\end{subfigure}
\begin{subfigure}{0.49\columnwidth}
\centering
\includegraphics[width=1\columnwidth]{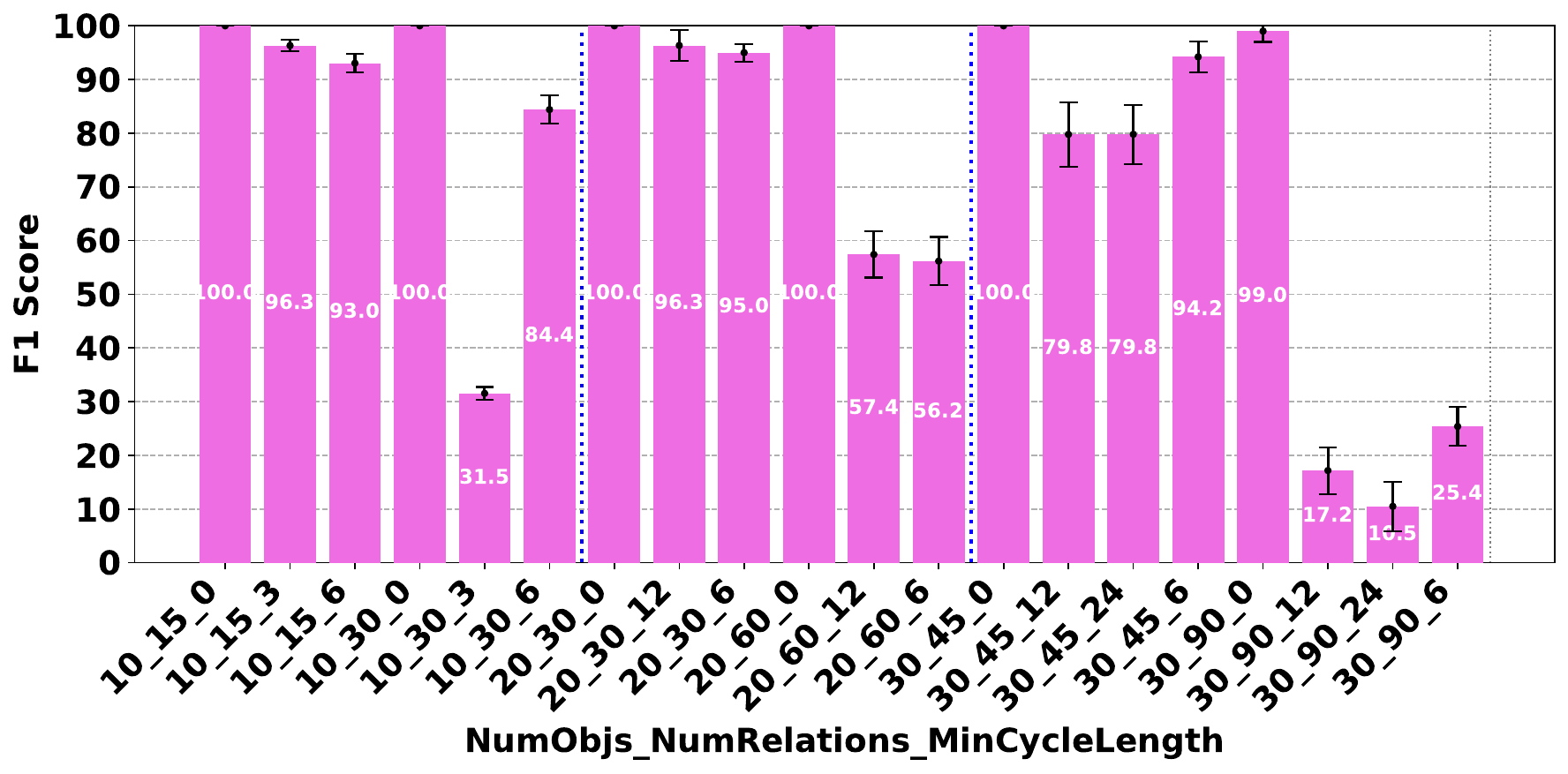}
\caption{O3-MINI.}
\label{fig:task2o3}
\end{subfigure}
\caption{Detailed Average F1 Score of \cons Task - Continued}
\label{fig:task2detail}
\end{figure*}

\autoref{fig:task2detail} shows the detailed result for \cons task. The average F1 score is aggregated based on the group of the number of objects, the number of relations and the minimum cycle length in the provided relations to the model. We performed 5 runs for each model, and calculated the average F1 score. 

\subsection{Detailed \qa Task Results}
\label{append:detailedEvalResulttask3}

\begin{figure*}[!t]
\centering
\begin{subfigure}{0.49\columnwidth}
\centering
\includegraphics[width=1\columnwidth]{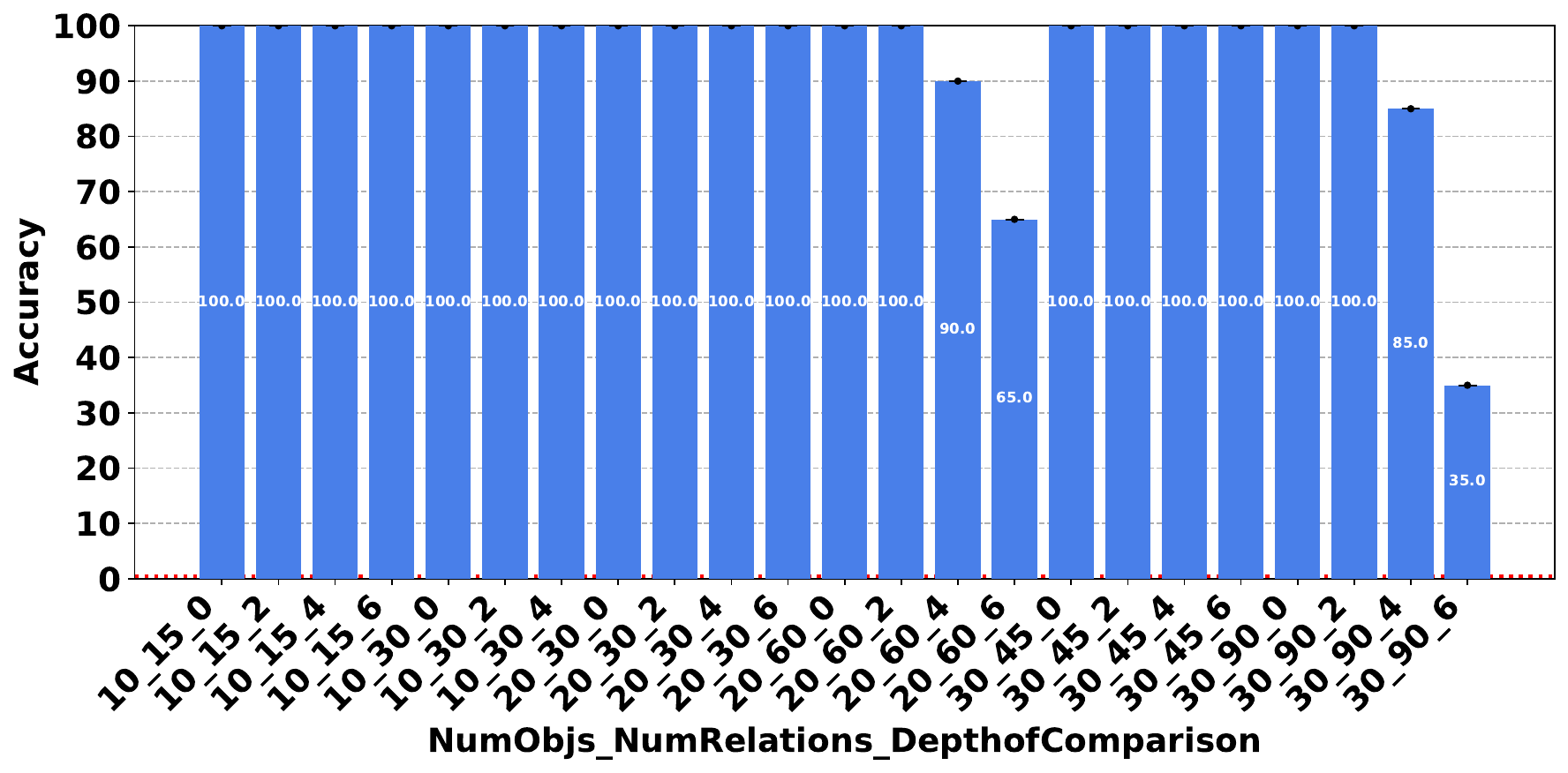}
\caption{Claude 3.7 Sonnet Thinking.}
\label{fig:task3Claude}
\end{subfigure}
\begin{subfigure}{0.49\columnwidth}
\centering
\includegraphics[width=1\columnwidth]{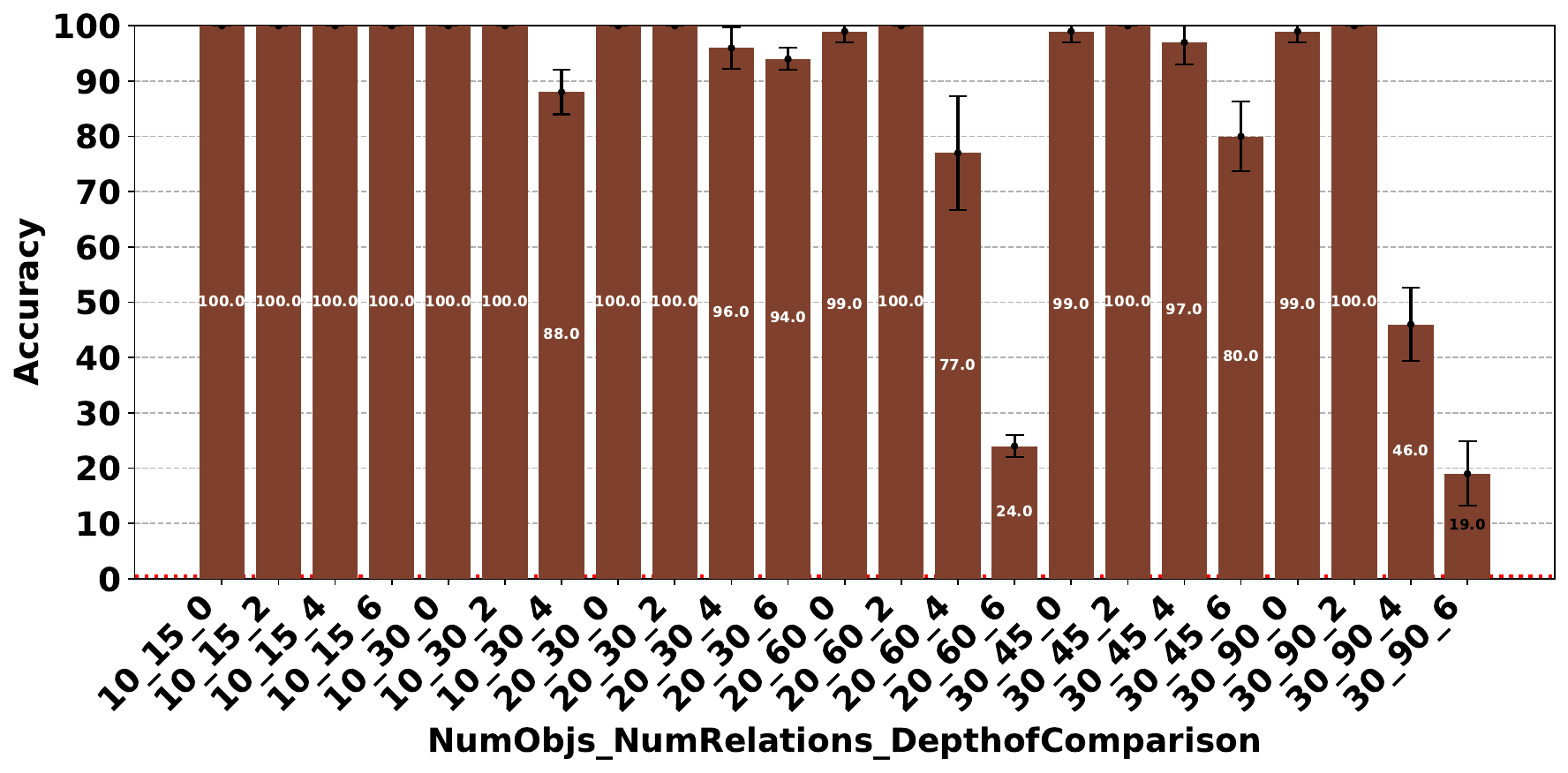}
\caption{Deepseek R1.}
\label{fig:task3deepseek}
\end{subfigure}
\begin{subfigure}{0.49\columnwidth}
\centering
\includegraphics[width=1\columnwidth]{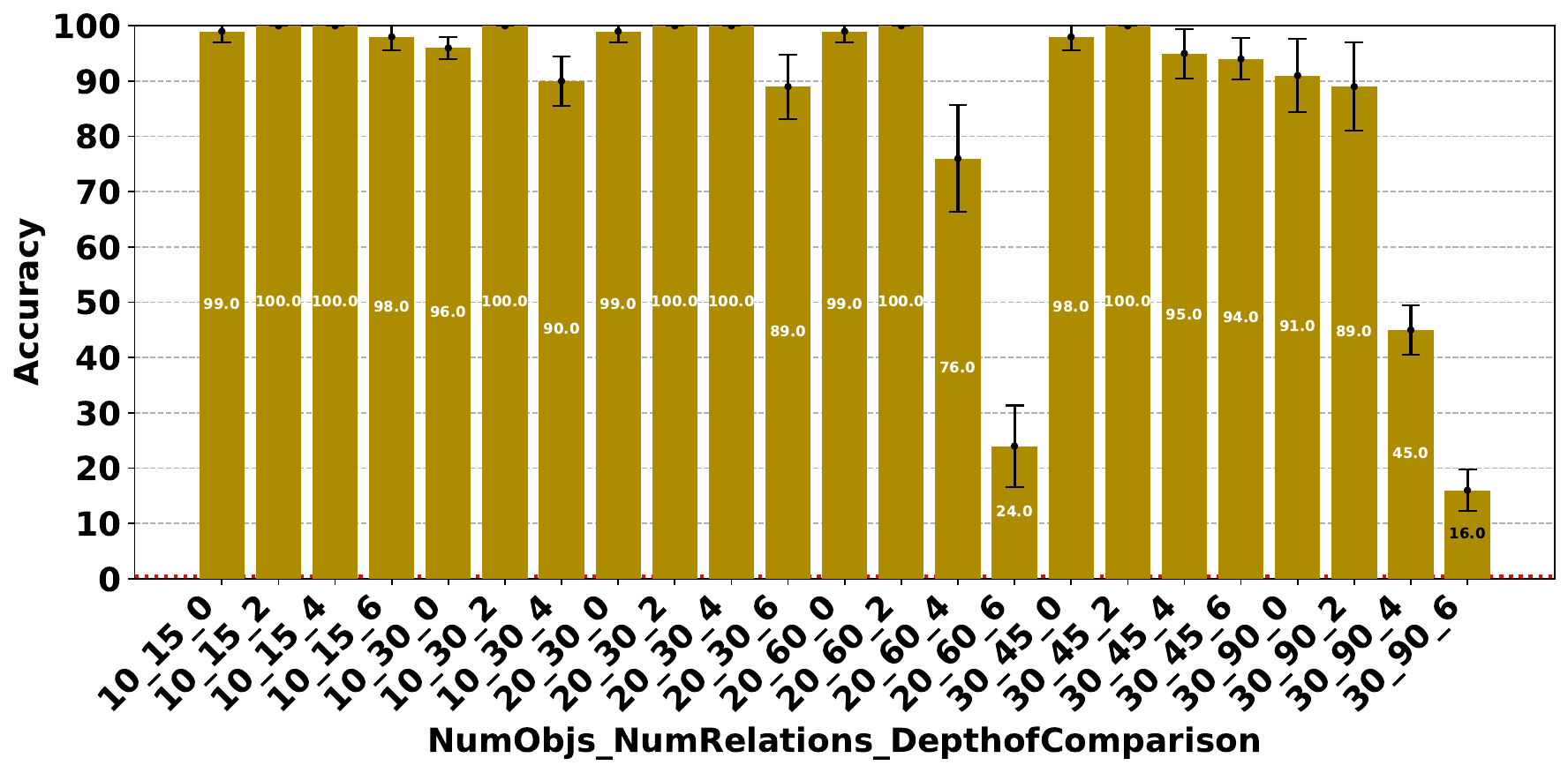}
\caption{Gemini 2 Flash Thinking.}
\label{fig:task3genimithinking}
\end{subfigure}
\begin{subfigure}{0.49\columnwidth}
\centering
\includegraphics[width=1\columnwidth]{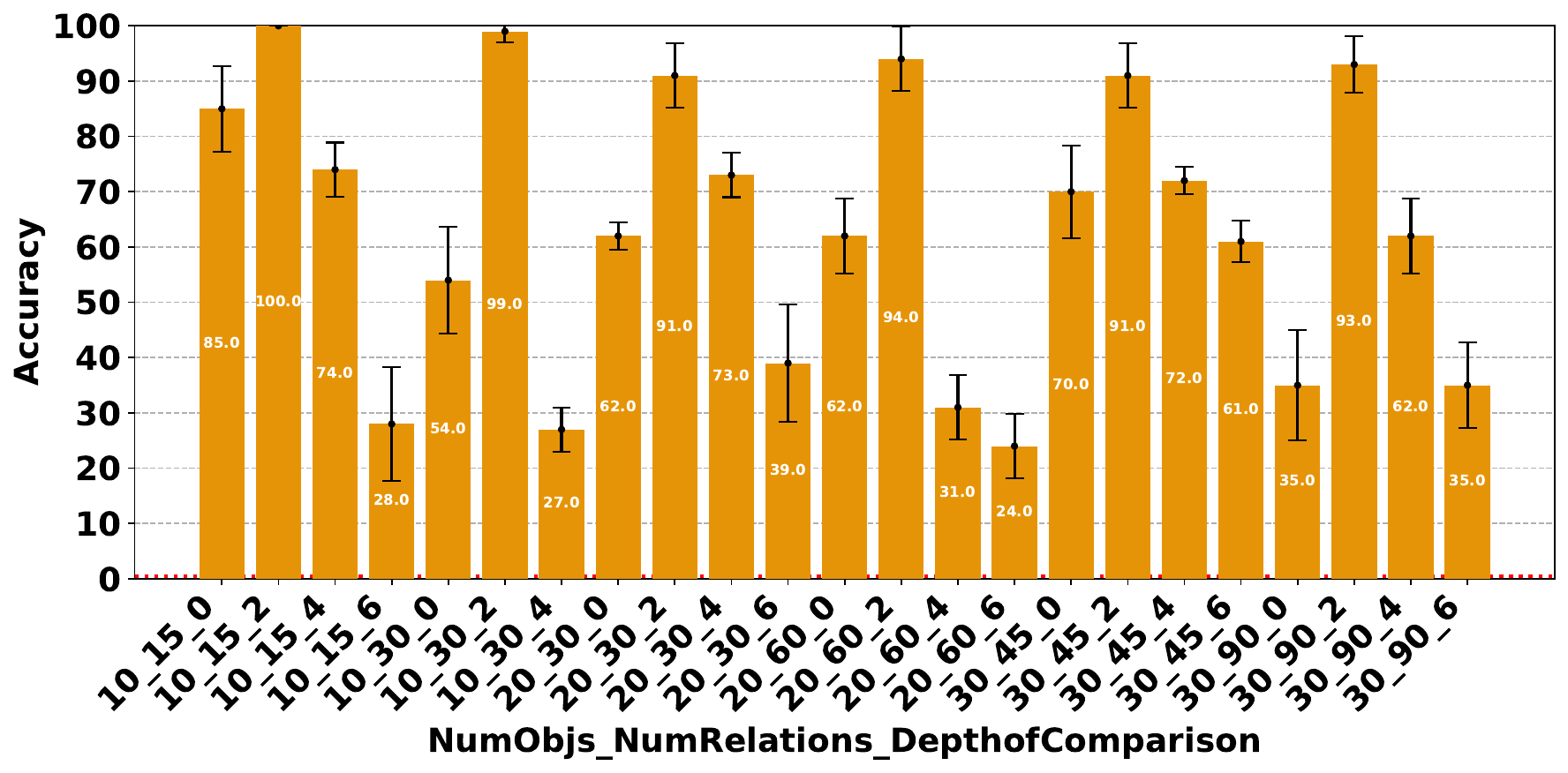}
\caption{Gemini 2.0 Pro.}
\label{fig:task3genimipro}
\end{subfigure}
\begin{subfigure}{0.49\columnwidth}
\centering
\includegraphics[width=1\columnwidth]{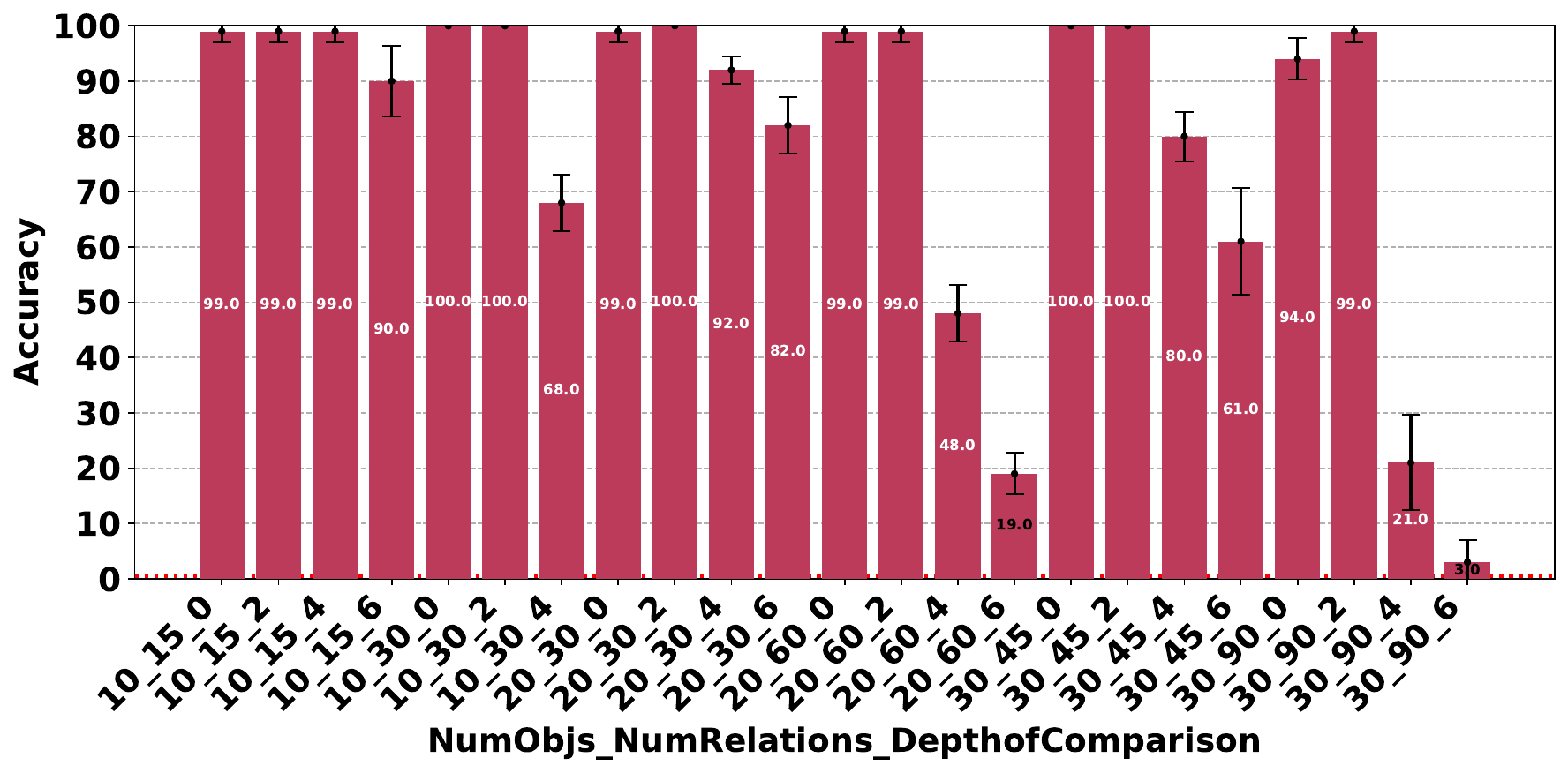}
\caption{GPT-4.5 Preview.}
\label{fig:task3gpt45}
\end{subfigure}
\centering
\begin{subfigure}{0.49\columnwidth}
\centering
\includegraphics[width=1\columnwidth]{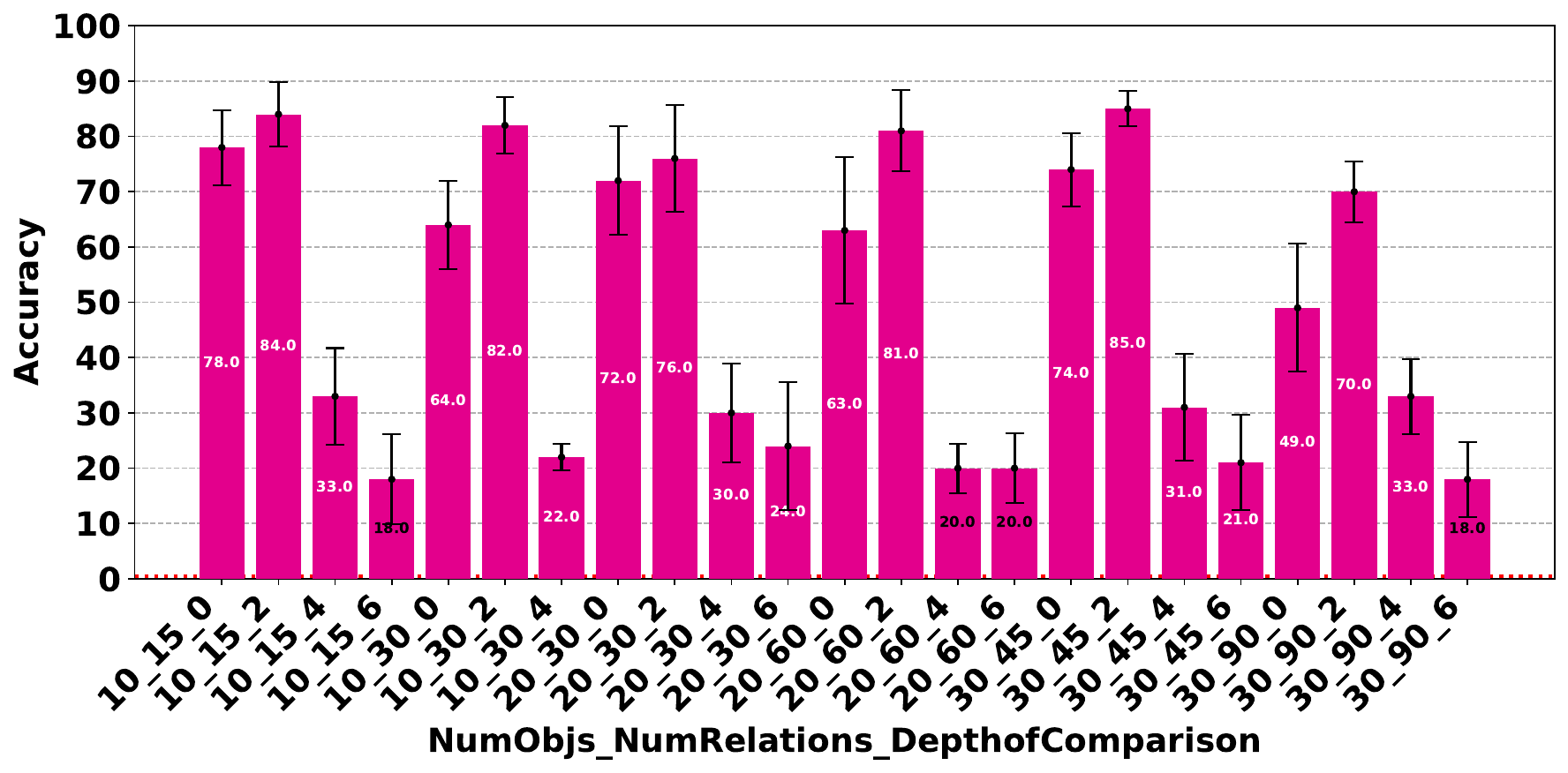}
\caption{GPT-4o.}
\label{fig:task3gpt4o}
\end{subfigure}
\begin{subfigure}{0.49\columnwidth}
\centering
\includegraphics[width=1\columnwidth]{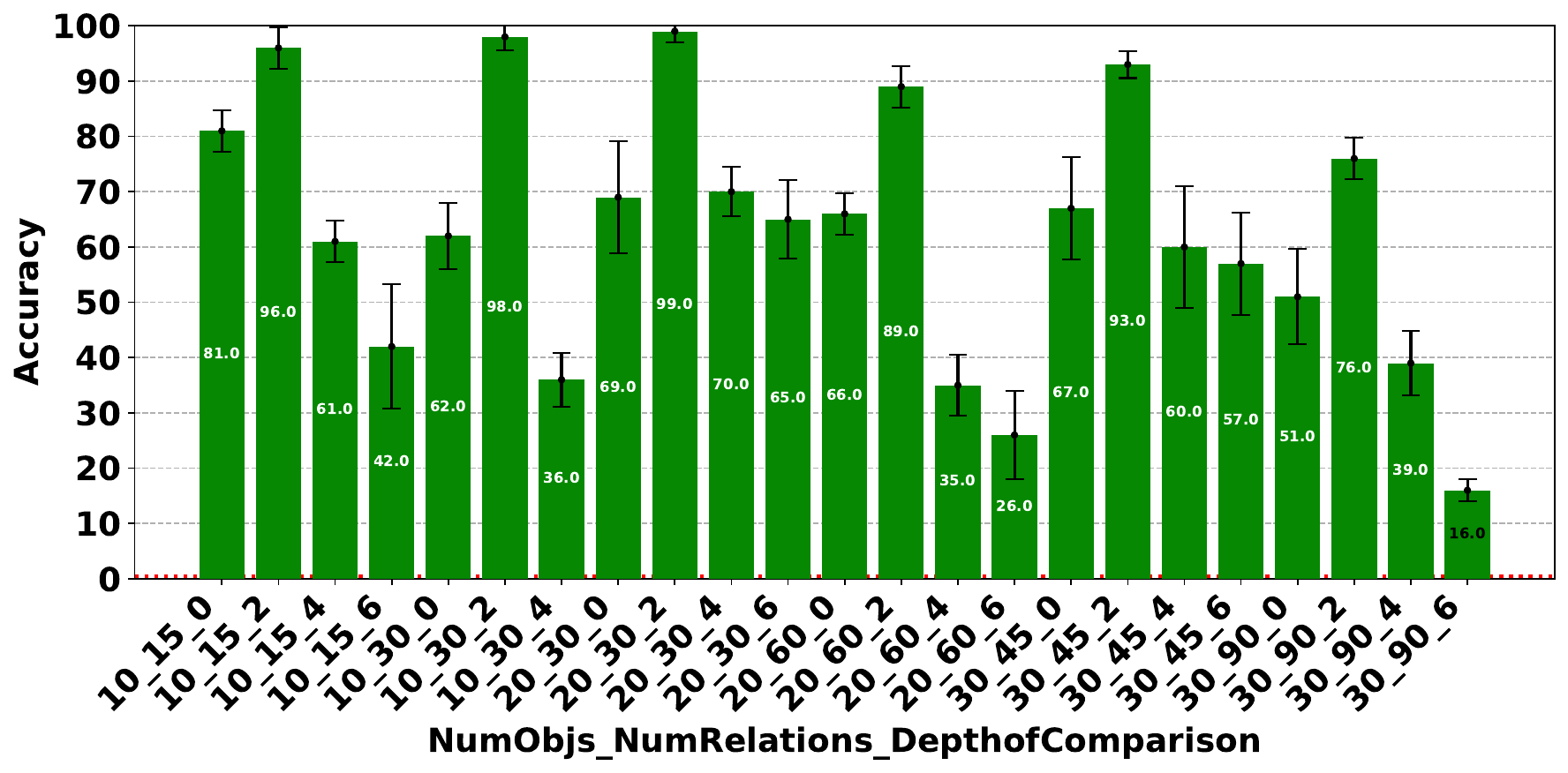}
\caption{Llama 3.1 405B Instruct.}
\label{fig:task3llama}
\end{subfigure}
\begin{subfigure}{0.49\columnwidth}
\centering
\includegraphics[width=1\columnwidth]{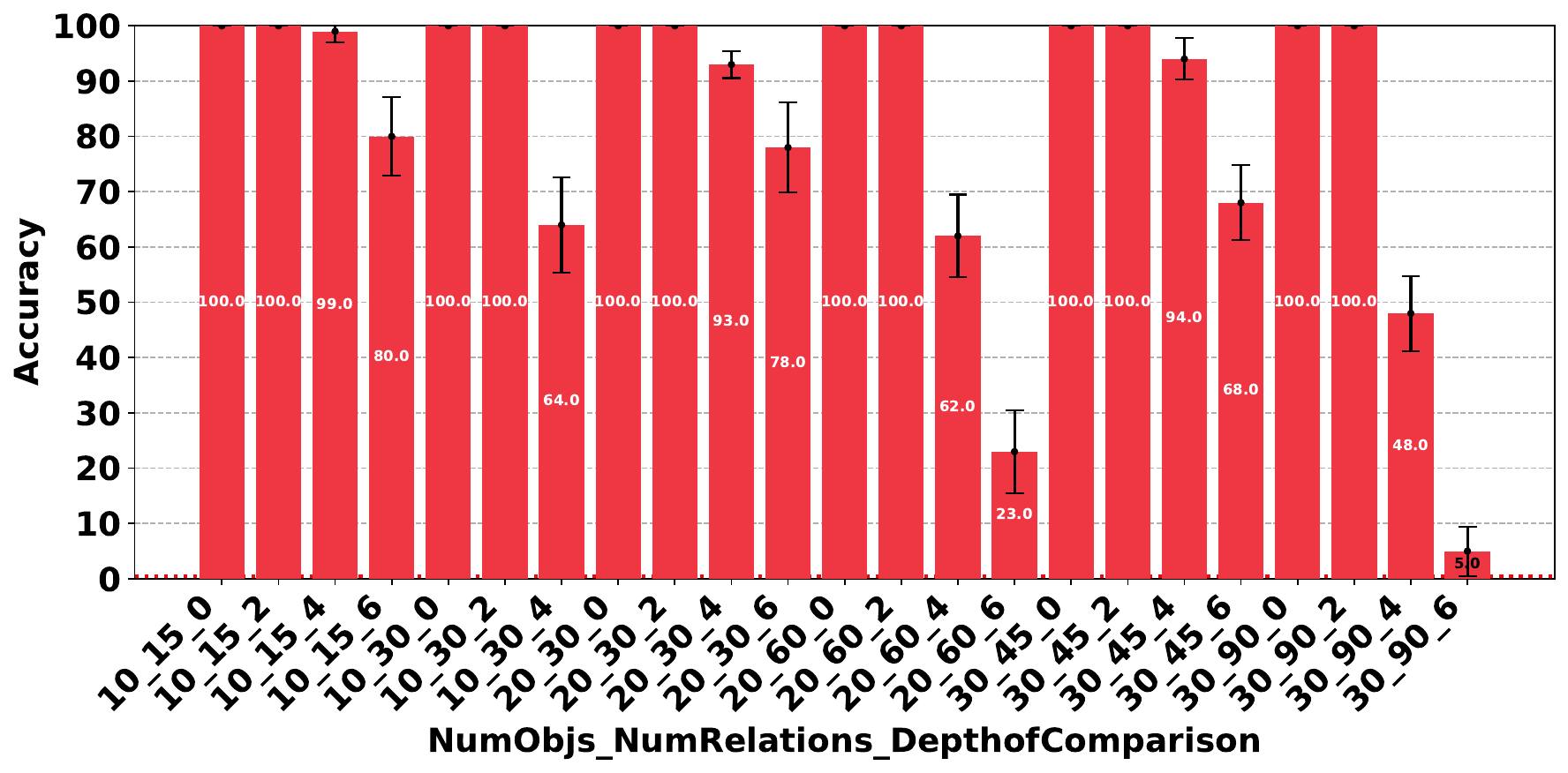}
\caption{O1.}
\label{fig:task3o1}
\end{subfigure}
\begin{subfigure}{0.49\columnwidth}
\centering
\includegraphics[width=1\columnwidth]{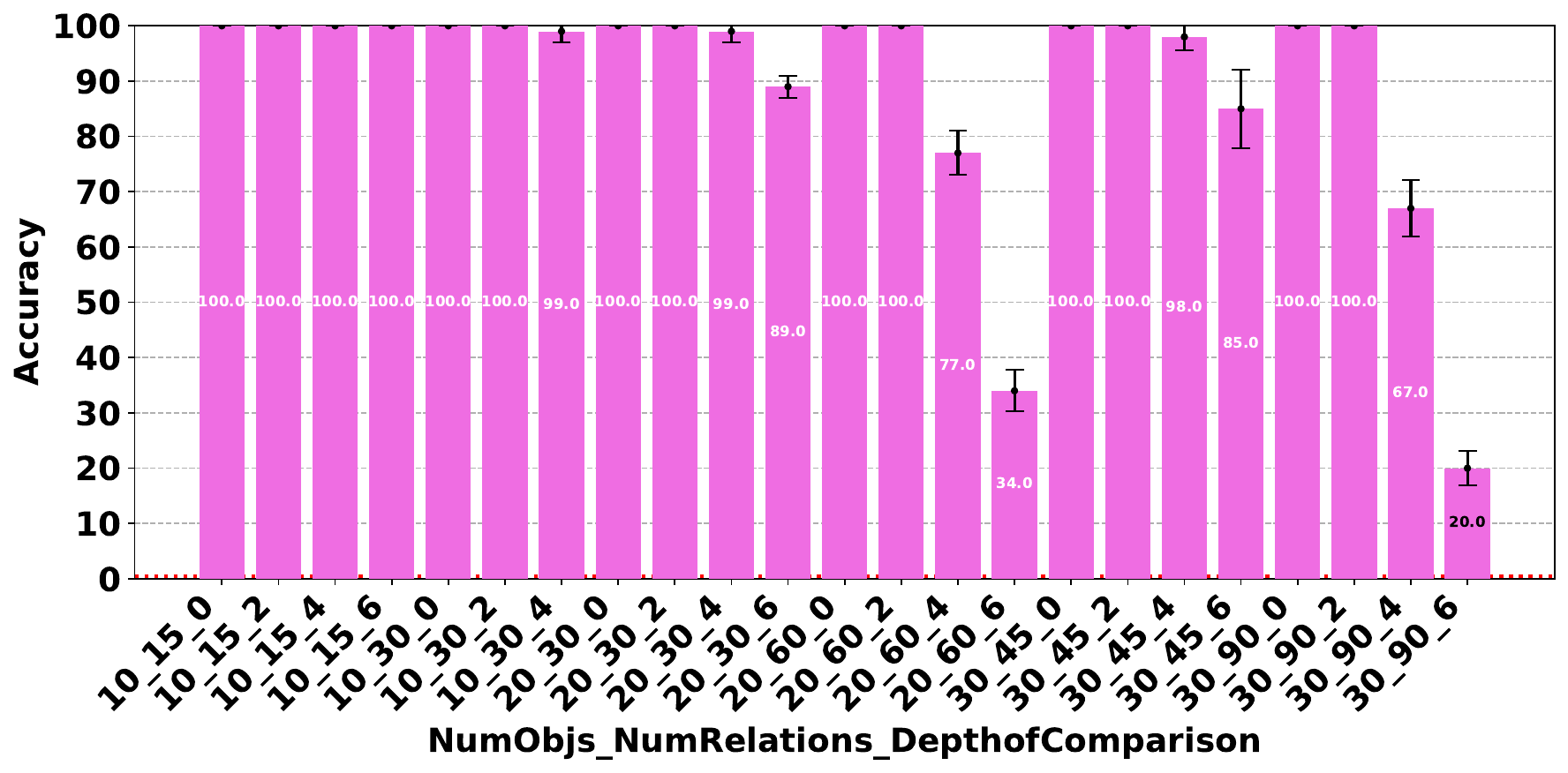}
\caption{O3-MINI.}
\label{fig:task3o3}
\end{subfigure}
\caption{Detailed Average Accuracy of \qa Task}
\label{fig:task3detail}
\end{figure*}

\autoref{fig:task3detail} shows the detailed result for \qa task. The average accuracy is aggregated based on the group of the number of objects, the number of relations and the depth of the provided comparison statement to the model. Except for \ClaudeSonnetThinking, We performed 5 runs for each model, and calculated the average. 

\end{document}